%% file: main.tex
\documentclass[11pt, a4paper,logo,nonumbering]{ms}
\usepackage[numbers, square, sort&compress]{natbib}
\usepackage{amssymb}
\usepackage{mathtools}
\usepackage{amsthm}
\usepackage{amsmath}
\usepackage[ruled,vlined,linesnumbered]{algorithm2e}
\usepackage{amssymb}
\usepackage{enumitem}
\usepackage{wrapfig}
\usepackage{url}
\usepackage{microtype}
\usepackage{dsfont}
\usepackage{multicol}
\usepackage{graphicx}
\usepackage{booktabs}
\usepackage{makecell}
\usepackage{enumitem}
\usepackage{listings}
\usepackage{wrapfig}
\usepackage{pifont}
\newcommand{\cmark}{\ding{51}}%
\newcommand{\xmark}{\ding{55}}%
\usepackage{tikz}
\usepackage{pgfplots}
\usepackage[export]{adjustbox}
\usepackage{tabularx}
\usepackage{tcolorbox}
\usepackage{fontawesome}
\usepackage[table]{xcolor}
\usepackage{hyperref}
\definecolor{citecolor}{HTML}{1976D2}
\hypersetup{
    colorlinks=true,
    linkcolor=citecolor,  % internal links incl. \ref now in citecolor
    citecolor=citecolor,
    urlcolor=cyan,
}

\makeatletter
\newcommand{\BlackTOC}{%
  \begingroup
    \renewcommand*{\@linkcolor}{black}%
    \renewcommand*{\@anchorcolor}{black}%
    \tableofcontents
  \endgroup
}
\makeatother

\usepackage{xspace}
\usepackage{pifont}% http://ctan.org/pkg/pifont
\usepackage{multirow}
\usepackage{tcolorbox}
\usepackage{xltabular}
\usepackage{longtable}
\usepackage{diagbox}
\usepackage{lineno}
\usepackage[bottom]{footmisc}
\usepackage{CJKutf8}
\usepackage{setspace}

\definecolor{cvprblue}{rgb}{0.21,0.49,0.74}
\definecolor{topcolor}{rgb}{1,0.8,0.8}
\definecolor{secondcolor}{rgb}{1,0.87,0.7}
\definecolor{thirdcolor}{rgb}{1,1,0.8}
\definecolor{promptcolor}{RGB}{200, 235, 227}
\definecolor{prompttitlecolor}{RGB}{175, 172, 172}

\mdfdefinestyle{prompt}{% 
    linecolor =   black,
    linewidth =   0.2pt,
    backgroundcolor =  promptcolor,
    roundcorner     =   1pt,
    font = \small\ttfamily,
    innerleftmargin=1.5pt,
    innerrightmargin=1.5pt
}

\newmdenv[
    roundcorner=5pt, 
    linecolor =   black,
    linewidth =   1pt,
    font = \small\ttfamily,
    subtitlebackgroundcolor=prompttitlecolor, 
    frametitlebackgroundcolor=prompttitlecolor,
    backgroundcolor=promptcolor, 
    frametitle={Generated caption},
    subtitleaboveskip=0.5\baselineskip,
    subtitlebelowskip=0.5\baselineskip,
    ]{captionenv}

\DeclareMathAlphabet{\pazocal}{OMS}{zplm}{m}{n}

\DeclareMathAlphabet\mathbfcal{OMS}{cmsy}{b}{n}

\makeatletter
\def\@BTrule[#1]{%
  \ifx\longtable\undefined
    \let\@BTswitch\@BTnormal
  \else\ifx\hline\LT@hline
    \nobreak
    \let\@BTswitch\@BLTrule
  \else
     \let\@BTswitch\@BTnormal
  \fi\fi
  \global\@thisrulewidth=#1\relax
  \ifnum\@thisruleclass=\tw@\vskip\@aboverulesep\else
  \ifnum\@lastruleclass=\z@\vskip\@aboverulesep\else
  \ifnum\@lastruleclass=\@ne\vskip\doublerulesep\fi\fi\fi
  \@BTswitch}
\makeatother

\addto\extrasenglish{
}

 {\begin{list}{}%
         {\setlength{\leftmargin}{#1}}%
         \item[]%
 }
 {\end{list}}
 
\reportnumber{001} % Leave blank if n/a
\renewcommand{\today}{}

\title{\centering Training-free Spatially Grounded Geometric Shape Encoding (Technical Report)}

\author{Yuhang He\\
\small Microsoft Research\\
}

\input{commands_tp.tex}

\input{secs_tp/sec0_abstract}

\begin{document}
\begin{CJK*}{UTF8}{gbsn}
\maketitle
\newpage
\begin{spacing}{0.9}
\BlackTOC
\end{spacing}
\newpage
\input{secs_tp/sec1_intro}
\input{secs_tp/sec2_relatedwork}
\input{secs_tp/sec3_method}
\input{secs_tp/sec4_exp}
\input{secs_tp/sec5_conclusion}
\input{secs_tp/sec6_appendix}
\newpage
\bibliography{main}
\newpage
\end{CJK*}
\end{document}

%% file: commands_tp.tex
\renewcommand{\phi}{\varphi}

\renewcommand{\leq}{\leqslant}

\renewcommand{\epsilon}{\varepsilon}
\renewcommand{\imath}{\mathrm{i}}

\newlength{\restsubwidth}
\newlength{\restsubheight}
\newlength{\restsubmoreheight}
\newcommand{\rest}[2]{%
        \settowidth{\restsubwidth}{\ensuremath{#2}}
        \settoheight{\restsubheight}{\ensuremath{{}_{#2}}}
        \ensuremath{{#1\hskip 0.5pt}_{\vrule\kern2pt\parbox[b][%
        4pt][b]{\the\restsubwidth}{%
                        \ensuremath{{}_{#2}}}}}
        }

%% file: secs_tp/sec0_abstract.tex
\begin{abstract}
Positional encoding has become the de facto standard for grounding deep neural networks on discrete point-wise positions, and it has achieved remarkable success in tasks where the input can be represented as a one-dimensional sequence. However, extending this concept to 2D spatial geometric shapes demands carefully designed encoding strategies that account not only for shape geometry and pose, but also for compatibility with neural network learning. In this work, we address these challenges by introducing a training-free, general-purpose encoding strategy, dubbed \textbf{XShapeEnc}, that encodes an arbitrary spatially grounded 2D geometric shape into a compact representation exhibiting five favorable properties, including invertibility, adaptivity, and frequency richness. Specifically, a 2D spatially grounded geometric shape is decomposed into its normalized geometry within the unit disk and its pose vector, where the pose is further transformed into a harmonic pose field that also lies within the unit disk. A set of orthogonal Zernike bases is constructed to encode shape geometry and pose either independently or jointly, followed by a frequency-propagation operation to introduce high-frequency content into the encoding. We demonstrate the theoretical validity, efficiency, discriminability, and applicability of \textbf{XShapeEnc} via extensive analysis and experiments across a wide range of shape-aware tasks and our self-curated \textit{XShapeCorpus}. We envision \textbf{XShapeEnc} as a foundational tool for research that goes beyond one-dimensional sequential data toward frontier 2D spatial intelligence.\\

\faEnvelopeO\ \hspace{0.2cm} Corresponding: \texttt{yuhanghe@microsoft.com}\\
\faGithub \hspace{0.2cm} Code: \url{https://github.com/yuhanghe01/XShapeEnc} \\
\end{abstract}

%% file: secs_tp/sec1_intro.tex
\section{Introduction}
\label{sec:intro}

Discrete positional encoding has been thoroughly investigated since the emergence of Transformer network architectures~\cite{roformer,att_all_need,hua2024fourierpe,shu20233DPPE,dosovitskiy2020vit,alibi}. It explicitly embeds discrete positions indexed by coordinates into a high-dimensional latent space, by which the follow-up deep neural network can be successfully grounded on the input discrete position. Serving as a standard practice, discrete positional encoding has been widely applied in position-sensitive tasks such as natural language processing~\cite{att_all_need}, image patch tokenization in computer vision~\cite{dosovitskiy2020vit}, microphone position encoding in acoustics~\cite{He_deepnerap_icml24} and point's coordinate encoding in 3D point clouds~\cite{shu20233DPPE}. Apart from point-wise position, 2D spatial geometric shapes often emerge as an integral object of interest that either deep neural networks or other methods need to reason about. Although there has been thorough discussion of discrete position encoding, research on unified 2D spatial geometric shape encoding has lagged far behind. \emph{How to encode an arbitrary 2D spatial geometric shape into a compact representation that is both structure-preserving, spatially sensitive and neural-friendly awaits thorough investigation.}

\begin{table}[h]
\centering
\small
\caption{Desirable properties comparison between potential 2D geometric shape encodings.}
\begin{tabular}{l|ccccc}
\hline
Method & \makecell[c]{\makecell[c]{arbitrary \\ shape?}} & \makecell[c]{\makecell[c]{high \\ frequency?}} & \makecell[c]{\makecell[c]{training-\\ free?}} & \makecell[c]{task-\\agnostic?} & \makecell[c]{spatial-\\ context ?}\\
\hline
AngularSweep~\cite{soundTRC,rezero} & \xmark & \xmark & \cmark & \xmark & \cmark\\
Poly2Vec~\cite{siampou2024poly2vec} & \xmark & \xmark & \xmark & \xmark & \cmark  \\
Space2Vec~\cite{space2vec_iclr2020} & \xmark & \xmark & \xmark & \xmark & \cmark \\
DeepSDF~\cite{deepSDF_2019_CVPR} & \cmark & \cmark & \xmark & \cmark & \xmark \\
2DPE~\cite{att_all_need} & \xmark & \xmark & \xmark & \cmark  & \cmark \\
ShapeEmbed~\cite{shapeembed} & \cmark & \cmark & \xmark & \cmark &\xmark \\
ShapeDist~\cite{shape_distribution} & \cmark & \xmark & \cmark & \cmark &\xmark \\
\hline
\emph{XShapeEnc}~(Ours) & \cmark & \cmark & \cmark & \cmark & \cmark \\
\hline
\end{tabular}
\label{table:encode_property_compare}
\end{table}

We conjecture that such under-exploration is due to two main reasons: the convenience of representing a 2D shape as a regular 2D image; the tight entanglement between 2D shape and task. First, influenced by the huge success of modern image-based deep neural networks~(\textit{e.g.}, CNNs~\cite{resnet18}, ViT~\cite{dosovitskiy2020vit} and generative models~\cite{yang2023reco}) and classic image-based feature extraction~\cite{contour_text_extract,lsd_linesegdet,contour_fragment}, 2D shapes are usually padded into regular 2D images and promising results can often be observed. Dedicated investigation of 2D geometric shape encoding has thus been largely inhibited. Second, existing works dealing with 2D shapes~\cite{soundTRC,rezero,siampou2024poly2vec,yu2024polygongnn} couple 2D geometric shape encoding with the task, in which they either focus on specific 2D geometric shapes~(\textit{e.g.}, polygon~\cite{siampou2024poly2vec,yu2024polygongnn}, sector shape~\cite{soundTRC,rezero,create_speech_zone}) or integrate the shape encoding into the task-aware, data-driven neural network learning process. The resulting shape encoding tuned on one task inevitably lacks transferability to other tasks. For example, a plethora of work~\cite{pointnet,shape_des_3D,shapenet2015} focus on the shape recognition task where the goal is to construct intra-category congruent, inter-category discriminative features, in which the shape's spatial position and intra-category structural difference have been intentionally ignored. The emergence of spatial intelligence~\cite{spatial_ai_davison} in recent years has brought 2D spatially grounded geometric shape encoding to the broader attention. Typical examples include spatial acoustic target region control~\cite{soundTRC,rezero,create_speech_zone} task, which tries to isolate spatial audio from a pre-specified spatial region, spatial region based text-to-image generation aims to place objects at text-specified position in the image~\cite{yang2023reco,make_a_scene}. As shown in Table~\ref{table:encode_property_compare}, all these methods vary drastically in geometric shape encoding~(whether spatially grounded or not), thereby hindering systematic benchmarking and limiting the progress of unified research in geometric shape modeling.

Motivated by the aforementioned discussion, we seek a unified 2D spatially grounded geometric shape encoding framework that exhibits five main desirable properties:

\begin{tcolorbox}[colframe=blue!40, colback=gray!5, title={XShapeEnc: Five Main Desirable Properties},boxrule=0.3mm, fontupper=\textit,]
1. \textbf{General-Purpose and Training-Free}: the encoding is task-agnostic and involves no training.\\
2. \textbf{Invertibility and Interpretability}: the encoding result is invertible in principle and interpretable.\\
3. \textbf{Controllability}: flexibly allows controllable balance between shape geometry and pose.\\
4. \textbf{Generality and Efficiency}: capable of encoding arbitrary 2D geometric shapes and is highly efficient.\\
5. \textbf{Adaptivity}: the encoding strategy can be flexibly tuned to fit various practical needs.
\label{colorbox1}
\end{tcolorbox}

In this work, we introduce \emph{XShapeEnc}, a framework that intrinsically preserves the aforementioned five desirable properties. Drawing inspiration from classical functional approximation theory and recent advances in frequency-rich positional encodings~\cite{roformer,att_all_need}, \emph{XShapeEnc} is built upon orthogonal Zernike basis~\cite{zernike_moment} defined over the unit disk, enabling the separate yet consistent encoding of both shape geometry and shape pose. Specifically, an arbitrary spatially grounded 2D geometric shape is decomposed into its normalized geometry within the unit disk and its spatial pose that specifies its position. By projecting the normalized shape geometry onto the orthogonal Zernike basis and optionally applying frequency propagation, we obtain a geometry encoding that is fully interpretable, efficient, and general-purpose. Meanwhile, the shape pose is represented as a harmonic pose field within the same unit disk so as to be encodable by Zernike basis. A controllable shape geometry and shape pose joint encoding strategy is introduced to enable adjustable emphasis on geometry or pose. The entire encoding process is task-agnostic, training-free and invertible in principle.

We demonstrate how \emph{XShapeEnc} fulfills the five desirable properties outlined above through: 1. rigorous mathematical derivation of the whole encoding process; 2. in-depth analysis on encoding discriminability regarding both shape geometry and pose encoding; 3. extensive experiments across various downstream tasks, including shape retrieval~\cite{shape_2d_dataset,latecki2006mpeg7}, pairwise shape topological relation classification~\cite{geospatial_big_data,siampou2024poly2vec} and target region control~\cite{soundTRC}. We envision \textbf{XShapeEnc} as a foundational tool for research beyond one-dimensional sequential data and data-driven, learning-based encoding paradigms, paving the way for a unified spatial encoding framework for frontier 2D spatial intelligence.

%% file: secs_tp/sec2_relatedwork.tex
\section{Related Work}

\subsection{Geometric Shape Related Task} 

Geometric shape-related tasks can be divided into two main categories depending on whether they require a shape's spatial context, where spatial context indicates a shape's spatial position, orientation, and scale. Most existing works on geometric shape focus on shape recognition~\cite{pointnet,shapenet2015,shape_des_3D,shapeembed,konukoglu2013wesd} in which the spatial context has been ignored. They tend to learn inter-class discriminative and intra-class consistent shape encoding. On the contrary, another line of work explicitly accounts for the shape's spatial context. If the spatial position, orientation, or scale of a geometric shape changes, its encoded shape representation changes accordingly. Such spatially grounded geometric shape encoding has been preliminarily explored in recent years within the broader context of spatial intelligence~\cite{spatial_ai_davison,soundTRC,rezero,yang2023reco,yu2024polygongnn,siampou2024poly2vec,zhang2025unitregionencodingunified,veer2019deeplearningclassificationtasks,mai2023towards}. However, these approaches are either tailored to specific tasks~\cite{soundTRC,rezero,create_speech_zone,yang2023reco} or designed for special geometric shapes~\cite{yu2024polygongnn,siampou2024poly2vec,zhang2025unitregionencodingunified,soundTRC,rezero,create_speech_zone}, intrinsically constraining the generalization and applicability of their shape encoding in handling arbitrary spatially grounded geometric shapes. Such diverse shape encoding requirements set by various tasks, as well as the intimate entanglement of 2D geometric shape encoding and specific tasks, result in a lack of thorough and independent investigation on geometric shape encoding. In this work, our proposed \emph{XShapeEnc} is capable of encoding arbitrary spatially grounded geometric shapes within a unified framework, and it can be further modified to cater to different encoding requirements.

\subsection{Training-Required Geometric Shape Encoding}

Most existing work dealing with 2D geometric shapes, whether they are spatially grounded or not, rely on deep neural networks to learn shape representation. Depending on if they entangle the learning process with the downstream task, they can be classified into two main categories. The methods in the first category learn shape representation in latent space for each shape separately~\cite{deepSDF_2019_CVPR,Occu_Net,atlasnet,octree_gennet_2017,chen2018implicit_decoder} or in a self-supervised manner~\cite{shapeembed}. Involving no task-centered training, they usually use shape itself as supervision signal. For example, the signed distance field~(SDF) family methods~(\textit{e.g.}, DeepSDF~\cite{deepSDF_2019_CVPR}, OccuNet~\cite{Occu_Net}) associate each individual shape with a learnable latent representation, which is further fed to a shared decoder network. The latent representation is optimized by predicting the shape decision boundary. Generative modeling methods~\cite{atlasnet,octree_gennet_2017,chen2018implicit_decoder} learn the shape representation by generating the geometric shape or shape surface. In recent years, autoencoder has been leveraged to learn shape representation in a self-supervised manner~\cite{shapeembed,O2VAE}. Methods in the second category~\cite{pointnet,shapenet2015,soundTRC,rezero} tightly couple the geometric shape encoding with the specific downstream task, in which the shape is implicitly learned and tailored for a specific task. These training-based encoding methods often require massive training datasets and are computationally expensive, their adopted neural networks need to be carefully designed to fit various training settings. On the contrary, our proposed \emph{XShapeEnc} is totally training-free and general-purpose. It provides shape encoding that is fully interpretable, invertible and frequency-rich, potentially serving as expressive representation for follow-up neural network learning.

\subsection{Training-Free Geometric Shape Encoding}

Training-free geometric shape encoding has a rich history in both image processing and statistical modeling domain. The prominent shape-relevant features such as curvature, perimeter and contour are widely used to represent a geometric shape~\cite{Chang1995ExtractingMS,contour_fragment,contour_text_extract,shape_distribution}. For example, we can use points sampled along a geometric shape's contour to represent a geometric shape. Distance matrix~\cite{shapeembed,O2VAE} or elliptical Fourier transform~(EFC~\cite{ellip_fourier_contour,ellip_fourier_shape} can be further applied on top of the contour points to extract more advanced shape representation. In addition to contour points, statistical modeling methods~\cite{shape_distribution} sample points across the whole shape area, then extract statistical feature from sampled points like pairwise distance. In parallel, decomposition methods are often used to approximate a 2D geometric shape by a set of bases. By projecting a 2D geometric shape onto each element of the basis, we can take the projection coefficients as the shape encoding. Typical decomposition includes elliptical Fourier transform~\cite{ellip_fourier_contour,ellip_fourier_shape}, Zernike basis~\cite{zernike_moment,zernike_image}, Legendre and Chebyshev polynomials. In this work, we rely on the Zernike basis for both shape geometry and shape pose encoding. As Zernike bases are mutually orthogonal and directly operate on shape area, \emph{XShapeEnc} naturally obtains compact shape representation and unifies shape geometry and shape pose encoding within the same framework.

Another line of recent work tends to integrate multiple training-free strategies or combine training-free and training-required encoding strategies together~\cite{soundTRC,rezero,siampou2024poly2vec,shapeembed,O2VAE,yu2024polygongnn} to encode a geometric shape. For example, based on the contour points, SoundTRC~\cite{soundTRC} applies positional encoding~\cite{att_all_need} to each point to get the shape encoding, ShapeEmbed~\cite{shapeembed} combines contour points extracted distance matrix and variational autoencoder~(VAE) to learn shape encoding, Poly2Vec~\cite{siampou2024poly2vec} combines Fourier transform and multilayer perceptron~(MLP). Despite their seemingly promising results, these methods impose strong assumptions on the type of geometric shapes they can encode~(\textit{e.g.}, the shape is convex~\cite{soundTRC,rezero}, simply connected without holes~\cite{soundTRC,rezero,shapeembed,O2VAE}, polygons~\cite{siampou2024poly2vec}). Such strong shape-type assumptions and sophisticated encoding strategies inhibit extending these encoding strategies to arbitrary shape and general-purpose geometric shape encoding. Our proposed \emph{XShapeEnc} overcomes all these obstacles, it unifies shape geometry and shape pose encoding within the same framework without requiring any training process. Extensive experimental results demonstrate the advantage of \emph{XShapeEnc}.

%% file: secs_tp/sec3_method.tex
\section{XShapeEnc: General 2D Geometric Shape Encoding}

\subsection{Problem Definition}

\begin{figure*}[t]
    \centering
    \includegraphics[width=0.99\linewidth]{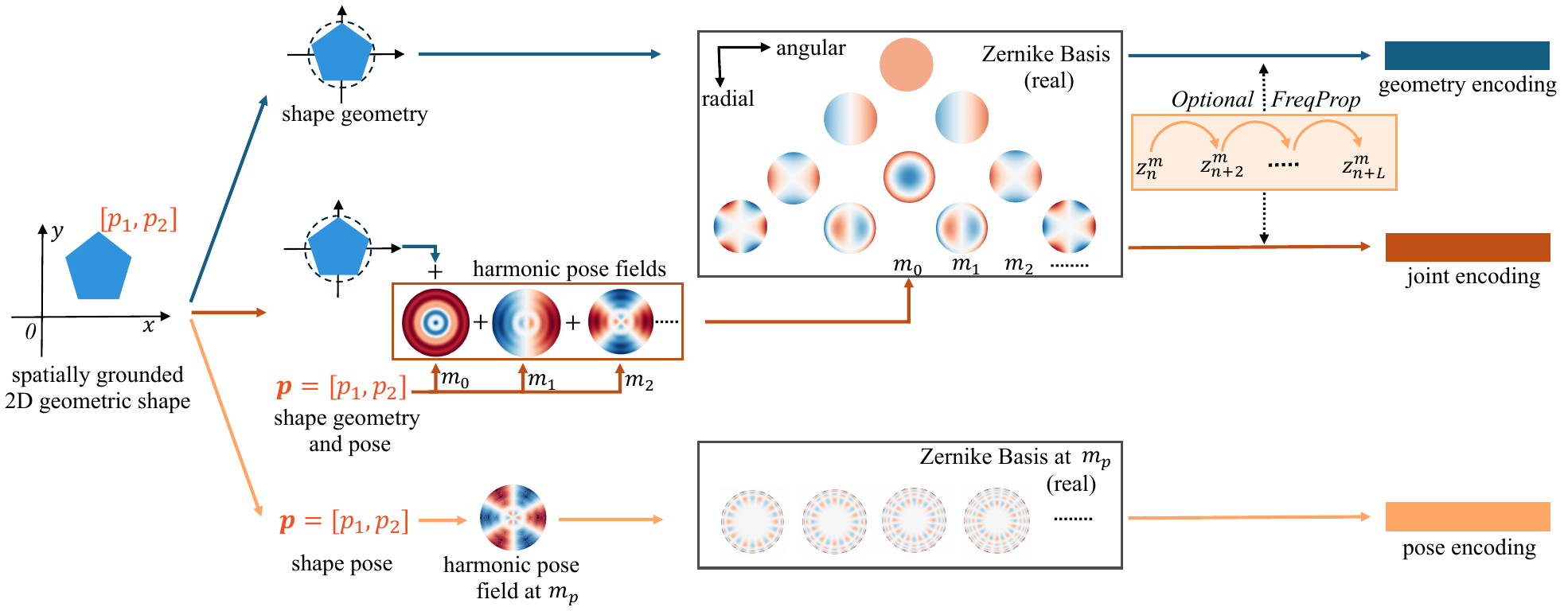}
    \vspace{-2mm}
    \caption{\emph{XShapeEnc} pipeline visualization. The spatially grounded shape is decomposed into its shape geometry within the unit disk and its shape pose vector. \emph{XShapeEnc} flexibly supports shape geometry encoding, shape pose encoding either independently or jointly, under the same Zernike basis umbrella. The shape pose vector constructs a harmonic pose field lying within the unit disk so that it can be processed by Zernike basis. In the shape geometry encoding and shape geometry and pose joint encoding, a set of Zernike basis across multiple radial and angular frequency bands are constructed, while in shape pose encoding a set of Zernike basis at a particular angular frequency band is constructed. A post hoc frequency propagation~(FreqProp) is optionally used to enrich the high-frequency in the encoding.}
    \label{fig:xshapeenc_pipeline}
    \vspace{-2mm}
\end{figure*}

We seek a general geometric shape encoding framework $\mathbfcal{F}$ that is capable of encoding an arbitrary spatially grounded 2D geometric shape~\footnote{In this work, the ``shape'' denotes a planar region with well-defined interior that admits a finite-resolution binary mask representation within a bounded domain, allowing digital rasterization and computational processing.} $S\subset\mathbb{R}^2$ into high-dimensional compact representation $Z \in \mathbb{R}^L$: $Z = \mathbfcal{F}(S)$. $\mathbfcal{F}$ exhibits desirable properties presented in the introduction section~\ref{sec:intro}. Since the 2D shape is spatially grounded, we decompose the shape $S$ into two orthogonal components: the normalized within-unit-disk shape geometry $S_{\mathcal{G}}$ that describes the shape's geometric structure~(\textit{e.g.}, contour, edge, etc.), and the shape pose $S_{\mathcal{P}}$ that describes the shape's spatial position~(\textit{e.g.}, scaling, translation, etc.), $S = [S_{\mathcal{G}}, S_{\mathcal{P}}]$. Specifically, the shape geometry $S_{\mathcal{G}}$ is represented by a 2D binary mask in the polar coordinate system $f_{\mathcal{G}}(r,\theta)$~($S_{\mathcal{G}} \coloneqq f_{\mathcal{G}}(r,\theta)$), while the shape pose consists of $2\times 2$ linear transformation matrix $A$ and a 2D translation vector $b$, $S_{\mathcal{P}}= [A, b]$. Throughout this decomposition, the input shape $S$ can be represented as $S = A\cdot S_{\mathcal{G}} + b$. By flattening the transformation $A$ and translation $b$ into a one-dimensional vector, we get the shape pose vector $\mathbf{p}\in \mathbb{R}^K$~($S_{\mathcal{P}} \coloneqq \mathbf{p}$). Taking the shape geometry $S_{\mathcal{G}}$ and shape pose $S_{\mathcal{P}}$ as input, \emph{XShapeEnc} is capable of encoding shape geometry and shape pose within a unified framework either independently or jointly, offering substantial encoding flexibility that can be further tailored for various practical needs~(see Fig.~\ref{fig:xshapeenc_pipeline} for the encoding pipeline visualization).

\begin{equation}
    Z = \{Z_{\mathcal{G}}, Z_{\mathcal{P}}, Z_{\mathcal{GP}}\};\, Z_{\mathcal{G}} = \mathbfcal{F}_{\mathcal{G}}(S_{\mathcal{G}}),\, Z_{\mathcal{P}} = \mathbfcal{F}_{\mathcal{P}}(S_{\mathcal{P}}), Z_{\mathcal{GP}} = \mathbfcal{F}_{\mathcal{GP}}(S_{\mathcal{G}}, S_{\mathcal{P}}); \mathbfcal{F} = \{\mathbfcal{F}_{\mathcal{G}}, \mathbfcal{F}_{\mathcal{P}}, \mathbfcal{F}_{\mathcal{GP}}\}
\label{eqn:problem_def}
\end{equation}

\noindent where $\mathbfcal{F}_{\mathcal{G}}$, $\mathbfcal{F}_{\mathcal{P}}$ and $\mathbfcal{F}_{\mathcal{GP}}$ are the encoding frameworks for shape geometry encoding, shape pose encoding and joint shape geometry and pose encoding, respectively. They are unanimously based on the unified Zernike basis transform framework.

\subsection{Zernike Basis Introduction}

To encode shape geometry and shape pose in Eqn.~(\ref{eqn:problem_def}) in a principled and compact representation, we adopt the Zernike basis as the encoding basis. Frits introduced by Frist Zernike in the 1930s~\cite{zernike_moment} for optical wavefront analysis, Zernike basis forms a set of complete and orthogonal basis over the unit disk and is defined in a polar coordinate system~$(r, \theta)$.

In the unit disk defined in polar coordinate system $\mathbb{D} = \{(r, \theta)\mid 0 \le r \le 1; 0 \le \theta \le 2\pi\}$, given a radial order $n$ and angular frequency band $m$ in $\mathbb{D}$, the complex Zernike basis $V_n^m(r, \theta)$ is the multiplication of radial polynomial $R_n^{|m|}(r)$ and angular frequency $e^{i m \theta}$. While the radial polynomial encodes the radial variation, the angular frequency captures the angular oscillation,

\begin{equation}
    V_n^m(r, \theta) = R_n^{|m|}(r) \cdot e^{i m \theta}
    \label{eqn:zernike_basis}
\end{equation}

\noindent where the radial polynomial $R_n^{|m|}(r)$ is defined as,

\begin{equation}
R_n^{|m|}(r)=\sum_{s=0}^{\frac{n-|m|}{2}}
(-1)^s
\frac{(n-s)!}{s!\left(\frac{n+|m|}{2}-s\right)!\left(\frac{n-|m|}{2}-s\right)!}r^{n-2s}
\end{equation}

\begin{wrapfigure}{r}{0.35\textwidth}
  \vspace{-4mm}
  \begin{center}
    \includegraphics[width=0.35\textwidth]{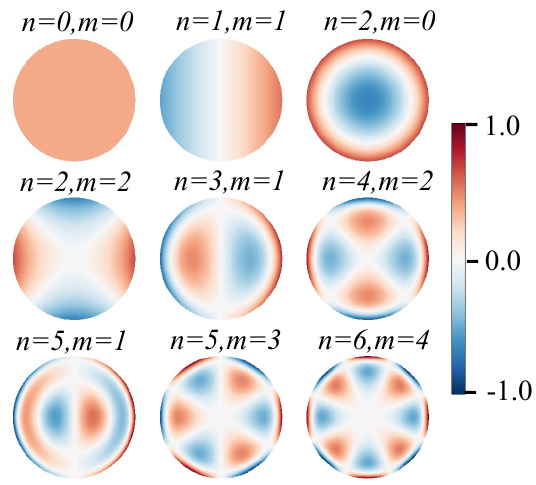}
  \end{center}
  \vspace{-4mm}
  \caption{Zernike basis visualization, we visualize nine real-part basis governed by radial mode $n$ and angular frequency $m$.}
  \label{fig:zernike_basis_vis}
\end{wrapfigure}

The Zernike basis $V_n^m(r, \theta)$ is in the complex domain $V_n^m(r, \theta)\in \mathbb{C}$, while $R_n^{|m|}(r)$ is in the real domain $R_n^{|m|}(r) \in \mathbb{R}$. By constraining $n - |m|$ to be even and $|m| < n$, the constructed Zernike bases are mutually orthogonal over the unit disk with respect to the area measure $r d_r d_\theta$, 

\begin{equation}
\int_{0}^{1}\!\!\int_{0}^{2\pi}
V_n^{m}(r,\theta)\,\big(V_{n'}^{m'}(r,\theta)\big)^{\!*}\,
r\,d\theta\,dr
=\frac{\pi}{n+1}\,\delta_{nn'}\delta_{mm'}
\label{eqn:zernike_basis_ortho}
\end{equation}

\noindent where $\delta_{mm'}$ and $\delta_{nn'}$ are Kronecker deltas. $\delta_{mm'} = 0$ if $m\neq m'$, otherwise $1$. $\delta_{nn'} = 0$ if $n\neq n'$, otherwise 1. The detailed mathematical proof is given in Sec.~\ref{app:sec:zernike_basis_ortho} in Appendix section. By selecting a set of radial order $n$ and angular frequency $m$ by following the $n - |m|$ is even and $|m| < n$ constraints, we can construct a Zernike basis set $\{V_n^m(r, \theta)| n = 0, 1, \cdots, N, m = -M, -M+2, \cdots, M\}$~(in Eqn.~(\ref{eqn:zernike_basis})). We visualize part of Zernike basis in Fig.~\ref{fig:zernike_basis_vis}, from which we can observe that the basis governed by different $n$ and $m$ exhibits different frequency variation along either the radial or angular direction. We exploit this characteristic to encode a 2D shape at different granularities.

Compared to conventional 2D Fourier transform, Zernike basis based transform exhibits several notable advantages: 1. its basis functions are polynomials rather than sinusoids, enabling more localized and compact shape representation; 2. the separation of radial and angular components allows independent control over radial resolution and angular frequency, facilitating flexible and fine-grained shape analysis; 3. Zernike basis based encoding possesses favorable properties including \textit{Linearity} which lends us the advantage to compositionally encode complex shape from simple shapes.

\subsection{Shape Geometry Encoding}
\label{sec:shape_geometry_encode}

Given the shape geometry $S_{\mathcal{G}}$ expressed as a binary mask in polar coordinate system: $f_{\mathcal{G}}(r,\theta)$, we project it onto the Zernike basis set $\{V_n^m\}$ and treat the projection coefficients as the shape geometry encoding. Specifically, for the Zernike basis $V_n^m(r,\theta)$, the projection coefficient $z_n^m$ is obtained by integrating the elementwise multiplication of the shape mask $f_{\mathcal{G}}(r,\theta)$ by $V_n^m(r,\theta)$ over the unit disk,

\begin{equation}
    z_n^m = \frac{n+1}{\pi} \iint_{\mathbb{D}} f_{\mathcal{G}}(r,\theta)\,[V_n^m(r,\theta)]^*\, r\, dr\, d\theta;\ \ z_n^m \in \mathbb{C}
\end{equation}

Different Zernike basis $V_n^m$ describes the shape geometry at different granularities: low-order coefficients describe the global outline, while higher-order coefficients encode localized and subtle details. By projecting onto Zernike basis, we obtain the shape geometry encoding $Z_{\mathcal{G}}$,

\begin{equation}
    Z_{\mathcal{G}} = [z_0^0,\, z_1^{-1},\, z_1^{1},\, z_2^{-2},\,\cdots,\, z_n^{m},\, \cdots,\, z_N^M ],\,\,\,\,\text{where}\,\,\,\, z_n^{m} \in \mathbb{C},\,\, m\le M, n\le N
    \label{eqn:shape_geometry_encode}
\end{equation}

\noindent where $M \le N,\ \  N-M \ \text{is even}$. The total number of complex coefficients in Eqn.~(\ref{eqn:shape_geometry_encode}) equals $\frac{N^2+M}{2} + 1$. Zernike basis based shape geometry encoding has two desirable properties that make the encoding in Eqn.~(\ref{eqn:shape_geometry_encode}) highly adaptive for various needs: 

\begin{enumerate}
    \item \textbf{Linearity}. Linearity means Zernike basis based encoding respects addition and scalar multiplication. The Zernike basis encoding on a composite shape geometry derived from adding and/or scaling different shape geometries together equals of the same linear operation on the Zernike basis encodings of each individual shape geometry. The mathematical proof is given in Sec.~\ref{sec:app:zernike_linearity_proof} in Appendix section. Benefiting from the \emph{Linearity} property, we can compositionally derive new shape geometry Zernike basis encoding by linearly compositing Zernike encodings of shape geometries that composite the new shape geometry, without having to encode the new shape geometry from scratch.
    \item \textbf{Rotation Equivariance}, rotating the input shape geometry equals to rotating the final Zernike basis, $Z_n^m(r,\theta + \phi) = Z_n^m(f)\cdot e^{-im\phi}$. Benefiting from the \emph{Rotation Equivariance} property, we can either derive rotation-invariant shape geometry encoding by taking the magnitude $Z_{\mathcal{G}}$, or rotation-variant feature by keeping the original encoding in Eqn.~\ref{eqn:shape_geometry_encode}.
\end{enumerate}

\subsection{High-Frequency Enrichment}
\label{sec:freqprop}

The encoding in Eqn.~(\ref{eqn:shape_geometry_encode}) is compact and invertible, but may suffer from high-frequency sparsity for simple shape geometries. For instance, the encoding for a circle only activates the lowest-order mode~(\textit{e.g.}, $Z_0^0$), resulting in the encoding dominated by $0$. Such low-frequency-only encoding lacks frequency richness and is suboptimal for neural network learning~\cite{highfreq_explain_cvpr2020,tancik2020fourfeat}. To resolve this challenge, we propose Frequency Propagation~(\emph{FreqProp}) that explicitly injects structural information informed by lower-frequency encoding to its adjacent higher-frequency encoding. Specifically, we introduce two kinds of frequency propagation strategies: radial frequency propagation~(\emph{rFreqProp}) and angular frequency propagation~(\emph{aFreqProp}). \emph{rFreqProp} enriches each coefficient $z_n^m$ by incorporating magnitude and phase information from its radial adjacent lower-frequency coefficient $Z_{n-2}^{m}$. \emph{aFreqProp} enriches each coefficient $z_n^m$ by incorporating magnitude and phase information from its angular adjacent lower-frequency coefficient $Z_{n}^{m-2}$,

\begin{equation}
    z_n^m \leftarrow z_n^m + \lambda_r |{z}_{n-2}^{m}|\cdot e^{i\arg(Z_{n-2}^{m})};\,\,\,\,z_n^m \leftarrow z_n^m + \lambda_a |{z}_{n}^{m-2}|\cdot e^{i\arg(Z_{n}^{m-2})}
\label{eqn:arfreqprop}
\end{equation}

\begin{wrapfigure}{r}{0.4\textwidth}
\centering
 \includegraphics[width=0.39\textwidth]{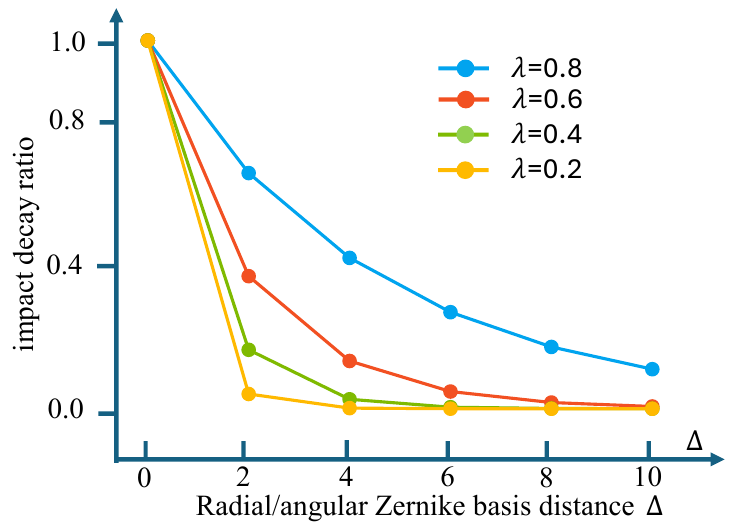}
  \caption{Frequency impact decay illustration in \textit{FreqProp}. We show the impact decay ratio w.r.t. radial/angular basis distance $\Delta_n$/$\Delta_m$, under the different propagation ratio $\lambda$.}
  \vspace{-2mm}
  \label{fig:impact_decay_vis}
\end{wrapfigure}
\begin{figure*}[t]
    \small
    \centering
    \begin{minipage}{0.48\linewidth}
        \centering
        \vspace{-5mm}
        \includegraphics[width=\linewidth]{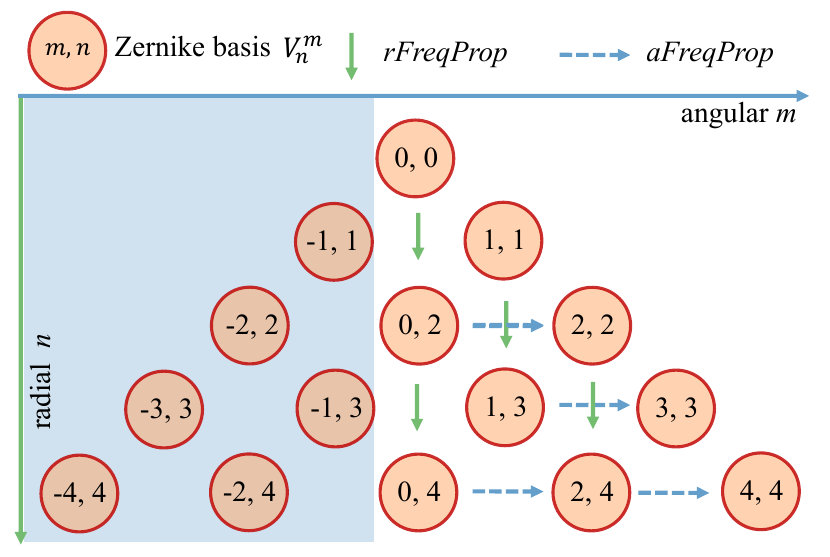}
        \vspace{-5mm}
        \caption{\textit{FreqProp} visualization. \textit{rFreqProp} and \textit{aFreqProp} propagate along fixed-angular and and fixed-radial Zernike basis, respectively. \textit{FreqProp} is invertible as we reverse the propagation process. We do not need to run \textit{FreqProp} on negative angular Zernike basis~($m<0$, overlaid by light blue) because their projection coefficients show conjugate symmetry~($z_n^{-m} = \overline{z_n^m}$) with their positive angular Zernike basis counterpart. We ignore the negative angular part when deriving the final encoding~(see Sec.~\ref{sec:enc_recap}).}
        \label{fig:freqprop_illu}
    \end{minipage}
    \hfill
    \begin{minipage}{0.48\linewidth}
        \centering
        \includegraphics[width=\linewidth]{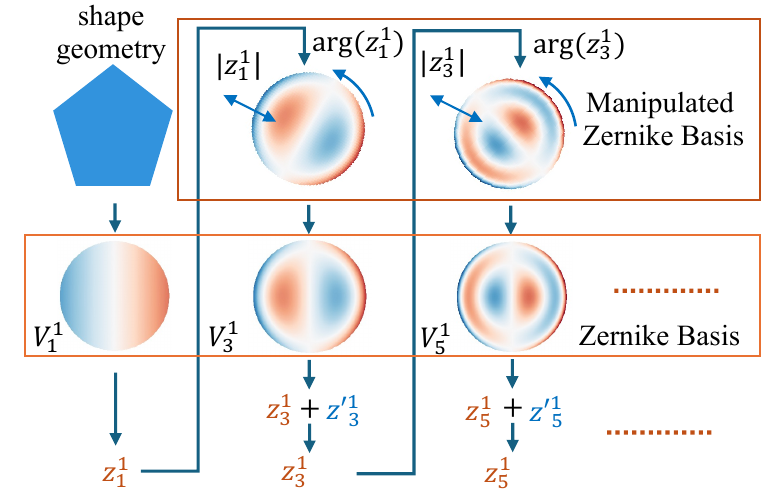}
        \vspace{-4mm}
        \caption{\emph{rFreqProp} geometric interpretation illustration. We use the angular frequency band $m=1$ as the propagation direction. The input shape geometry is projected onto the Zernike basis $V_1^1$ to get the projection coefficient $z_1^1$, which is then used to manipulate the next adjacent Zernike basis $V_3^1$ by scaling it by $|z_1^1|$ and rotating it by $\arg(z_1^1)$ before projecting onto $V_3^1$. The same rule applies to all other follow-up Zernike bases, including $Z_5^1$, $Z_7^1$, \textit{etc}.}
        \label{fig:freqprop_vis}
    \end{minipage}
    \vspace{-2mm}
\end{figure*}


\noindent where $\arg(\cdot)$ indicates the phase information, $\lambda_r$ is the radial propagation ratio deciding the frequency ratio propagating from lower-frequency coefficient~(we set it as 0.6, see Sec.~\ref{sec:app:prop_ratio_choice} in Appendix), $\lambda_a$ is the corresponding angular propagation ratio. As Zernike indices obey parity, stepping $n$ by 2~(or $m$ by 2) guarantees to land on valid Zernike index. Starting from the lower-frequency coefficient, we iteratively propagate the its frequency to the radial or angular higher-frequency coefficient by the chain rule in Eqn.~(\ref{eqn:arfreqprop}). The propagation ratio ensures the long-range exponential frequency influence decay. As is shown in Fig.~\ref{fig:impact_decay_vis}, a coefficient mostly impacts its adjacent higher frequency coefficient, the longer-range leads to exponentially decayed impact. This long range decay protects the encoding from contamination by frequency propagation and maintains the encoding discriminability. The whole propagation process is shown in Fig.~\ref{fig:freqprop_illu}.

Our proposed \textit{FreqProp} in Eqn.~(\ref{eqn:arfreqprop}) has clear geometric interpretation. As is shown in Fig.~\ref{fig:freqprop_vis}, it can be viewed as we perturb the higher-order Zernike basis $Z_n^m$ by \textit{rotating} it by $\arg(Z_{n-2}^{m})$ and \textit{scaling} it by $Z_{n}^{m}$. The perturbed basis is then ``projected'' back to itself to yield a non-zero coefficient as long as the lower-frequency is non-zero. Benefiting from the \textit{Linearity} and \textit{Rotation Equivariance} property of Zernike transform, we can directly obtain the propagated coefficient without rotating/scaling the Zernike basis in reality. Given the initial encoding in Eqn.~(\ref{eqn:shape_geometry_encode}). It fully accommodates the shape geometry's specificity as the propagation builds on top of the initial encoding. \emph{FreqProp} naturally preserves the orthogonality of the Zernike basis as no Zernike basis is altered during the propagation process.

Furthermore, the shape geometry encoding with radial frequency propagation still satisfies the \textit{Linearity} property, for which the mathematical proof is provided in Sec.~\ref{app:rfreqprop_linearity_proof} in Appendix. With the \textit{Linearity} property, we can easily derive complex shape geometry's encoding by linearly combine simple shape geometries' encoding.

\textbf{Invertibility}. \emph{FreqProp} is fully invertible, the original encoding coefficients can be precisely recovered by subtracting the propagated term, where the propagation term can be deterministically decided by starting from $Z_0^0$. With the recovered original coefficients, we can further recover the original shape geometry by $S_{\mathcal{G}} = \sum_{(n,m)\in (N, M)} Z_n^m \cdot V_n^m$.

\subsection{Shape Pose Encoding}
\label{sec:shape_pose_enc}

For the pose vector $\mathbf{p} \in \mathbb{R}^K$, we seek a pose encoding strategy that packs multiple scalar ``pose'' parameters into compact representation under the same Zernike basis. To this end, we define a harmonic pose field $f_{\mathcal{P}}(r, \theta; m_p, \mathbf{p})$ over the unit disk $(r, \theta)$ at the angular frequency $m_p$,

\begin{equation}
    f_{\mathcal{P}}(r, \theta; m_p, \mathbf{p}) = \Bigl(\sum_{k=1}^K p_k w_k(r)\Bigr) \cos(m_p \theta)
    \label{eqn:harmonic_pose_field}
\end{equation}

\noindent where $p_k \in [-1, 1],\ \  p_k \in \mathbf{p}$. For the sake of numerical stability, $\mathbf{p}$ is $L_2$-normalized. Each pose parameter $p_k$ is associated with a separate radial window $w_k(r)$, which defines how to place pose parameter energy along the radial axis. $\cos(m_p\theta)$ is the angular harmonic, we directly instantiate it as the real part of Zernike basis $e^{im\theta}$~(the imaginary part is ignored). By projecting the harmonic pose field onto Zernike basis $V_n^m(r, \theta)$, we only get non-zero projection coefficients $a_n^m = \langle f_{\mathcal{P}}(r, \theta, m_p), V_n^m(r, \theta)\rangle$ at Zernike basis whose angular frequency equals to $m_p$~(The detailed proof is presented in Sec.~\ref{sec:app:harmonic_posefield_derive} in Appendix),

\begin{equation}
    a_n^m = 
    \begin{cases}
        \displaystyle
        \pi \sqrt{\tfrac{n+1}{\pi}} \sum\limits_{k=1}^K p_k \int_0^1 w_k(r)\, V_n^{|m|}(r)\, r\, dr, & m = \pm m_p, \\
        0, & \text{otherwise}.
    \end{cases}
    \label{eqn:harmonic_field_coeff}
\end{equation}

From Eqn.~(\ref{eqn:harmonic_field_coeff}), we can learn that the projection coefficient falls exactly into the specific Zernike $(n, m_p)$ mode. The resulting coefficient vector $\{a_n^{\pm m_p}\}_{n\ge |m_p|}$ encodes the full pose vector $p$ in a Zernike-compatible and invertible manner. Let $C_{n,m_p}^k = {\textstyle \int_0^1} w_k(r)\, R_n^{|m_p|}(r)\, r\, dr$, we stack $C_{n,m_p}^k$ to get $\mathbf{C}$, stack $a_n^{m_p}$ to get $\mathbf{A}$, Eqn.~(\ref{eqn:harmonic_field_coeff}) can be rewritten as,

\begin{equation}
    \mathbf{A} = \mathbf{p} \cdot \mathbf{C},\ \ \text{where}\ \  \mathbf{A} \in \mathbb{R}^{1\times L}, \mathbf{p} \in \mathbb{R}^{1\times K}, C \in \mathbf{R}^{K\times L}
\label{eqn:pose_encode_matrix}
\end{equation}

\begin{wrapfigure}{r}{0.35\textwidth}
  \begin{center}
        \includegraphics[width=0.35\textwidth]{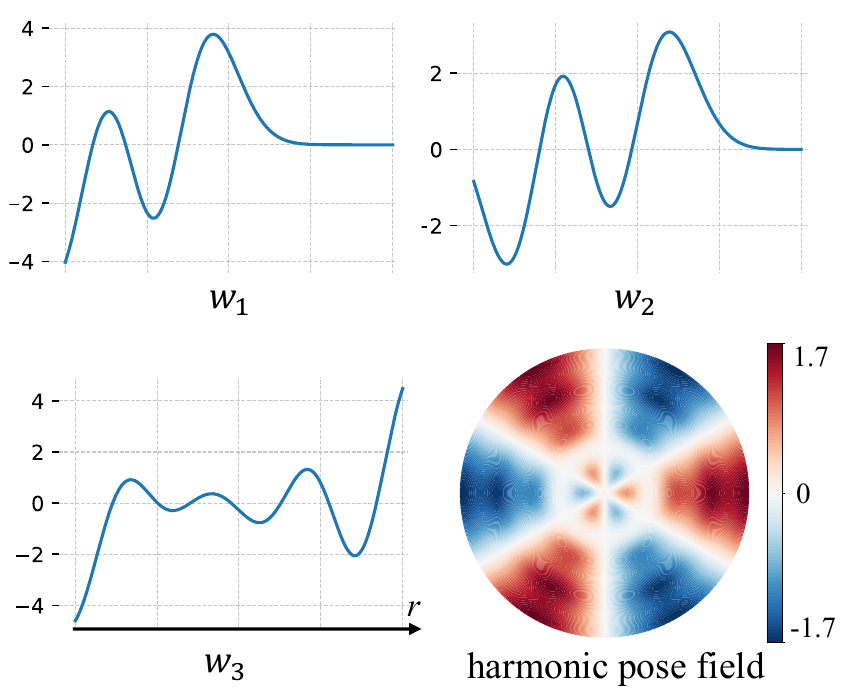}
  \end{center}
  \vspace{-4mm}
  \caption{Three orthonormal radial windows and harmonic pose field visualization.}
  \label{fig:pose_field_vis}
  \vspace{-3mm}
\end{wrapfigure}

To ensure invertibility and robustness of $\textbf{p}$, $\mathbf{C}$ has to be full rank \texttt{rank}($\mathbf{C}$)$=K$ and well-conditioned so that $\mathbf{p} = \mathbf{A}\mathbf{C}^{-1}$. To this end, we add two constraints to Eqn.~(\ref{eqn:harmonic_pose_field}): First, $K\le L$, which is the prerequisite for $\mathbf{C}$ to be full-ranked. In practice, we simply need to project the harmonic pose field to at least $K$ Zernike basis. Second, we instantiate $w_k(r)$ to be a set of \emph{radially
orthonormal} windows under the Zernike weight $r\,dr$ across different pose parameters~(see Fig.~\ref{fig:pose_field_vis}), with which each pose parameter $p_k$ only contributes to one distinct direction in the radial space, $ \langle w_i, w_j \rangle = \int_{0}^{1}w_i(r)w_j(r)\,r\,d_{r} = \delta_{ij}$. Radially orthogonal windows make $\mathbf{C}$ well-conditioned and often diagonal after projection, thus ensure the invertibility of $\mathbf{p}$. To explicitly introduce both low and high frequency along the radial direction, we implement each radial window $w_k$ to be two Gaussian bumps. The mutual radial orthonormal property are ensured by Gram-Schmidt based radial window generation. We show three constructed radial window $w$ as well as the harmonic pose field in Fig.~\ref{fig:pose_field_vis}. With the harmonic pose field in Eqn.~(\ref{eqn:harmonic_pose_field}), we get the shape pose encoding as,

\begin{equation}
    Z_{\mathcal{P}} = [a_{n}^{m_p},\  a_{n+2}^{m_p},\  a_{n+4}^{m_p},\  \cdots,\  a_N^{m_p}],\ \ a_n^{m_p} \in \mathbb{R}
    \label{eqn:pose_encode_vec}
\end{equation}

\noindent where $n = |m_p|,\,\, n < N,\ \  N-n \ \text{is even}$. The real-valued encoding length is $L=\frac{N-n}{2}$.

\begin{wrapfigure}{r}{0.35\textwidth}
\vspace{-4mm}
  \begin{center}
        \includegraphics[width=0.35\textwidth]{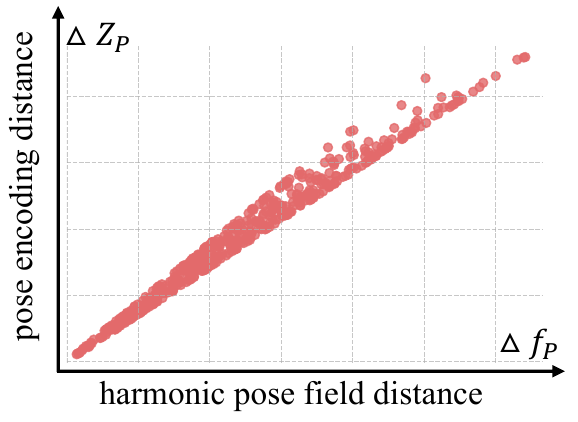}
  \end{center}
  \vspace{-4mm}
  \caption{Correlation between harmonic pose filed and the final shape pose encodings.}
  \label{fig:pose_field_corr}
  \vspace{-3mm}
\end{wrapfigure}

To verify that the proposed harmonic pose encoding is meaningful, we analyze how distances in the latent pose space correlate with the actual spatial effect of pose on the shape. For two pose vectors $\mathbf p_i$ and $\mathbf p_j$, we measure (i) their distance in the learned pose embedding $\|A(\mathbf p_i)-A(\mathbf p_j)\|_2$, and (ii) the $L^2$ distance between the corresponding pose fields $\|f_{\mathcal P}(\cdot;\mathbf p_i)-f_{\mathcal P}(\cdot;\mathbf p_j)\|_{L^2(\mathbb D)}$, which quantifies how much the shape is displaced in the image domain. As shown in Fig.~\ref{fig:pose_field_corr}, these two quantities exhibit strong linear correlation across randomly sampled pose pairs. This behavior is not incidental: because our pose field construction and Zernike projection are both linear, the pose embedding constitutes an isometric mapping of the pose--field Hilbert space, and distances in the latent space exactly reflect differences in the induced spatial fields. Importantly, this implies that the encoding does not preserve Euclidean distances in raw pose--parameter space, but rather preserves the \emph{functional effect of pose on the shape}, which is the quantity of interest for shape--pose reasoning. The observed correlation therefore confirms that our representation provides a geometrically meaningful and physically grounded embedding of pose.

\textbf{Linearity:} The proposed harmonic pose encoding is linear with respect to the pose field and the subsequent Zernike projection~(see the Proof in Sec.~\ref{sec:app_pose_linear}). In particular, the resulting pose coefficients are linear functions of the pose vector, and therefore obey superposition. However, we do not enforce linearity with respect to physical pose parameters in Euclidean space (e.g., translation), as such transformations are inherently non-linear in any fixed orthogonal basis on a bounded domain. Instead, our formulation embeds pose as a band-limited harmonic signal aligned with the Zernike basis, yielding a representation that is linear, controllable, and compatible with joint geometry–pose encoding.

\textbf{Pose Invertibility.} Once we ensure $K\le L$ and $w_k(r)$ is instantiated as a set of radially orthonormal windows, the original pose vector can be precisely recovered from pose encoding by Eqn.~(\ref{eqn:pose_encode_matrix}).

\subsection{Shape Geometry and Pose Joint Encoding}
\label{sec:joint_encode}

We present shape geometry encoding in Sec.~\ref{sec:shape_geometry_encode} and shape pose encoding in Sec.~\ref{sec:shape_pose_enc} separately. Although both relying on Zernike basis, they require to construct different Zernike basis for encoding. While shape pose encoding requiring to construct the Zernike basis at predefined fixed angular frequency band, shape geometry encoding instead require to construct the Zernike basis across multiple angular frequencies. To jointly encode shape geometry and pose into one representation $Z_{\mathcal{GP}}$ and with exactly the same Zernike basis, we propose a novel shape geometry and shape pose joint encoding framework.

Two most straightforward approaches are to jointly encode in the final feature space via either \textit{Addition} or \textit{Concatenation}. For \textit{Addition}, the geometry encoding $Z_{\mathcal{G}}$ is combined with the pose encoding $Z_{\mathcal{P}}$ through elementwise addition, weighted by a scalar $\alpha$. For \textit{Concatenation}, the two encodings are simply concatenated:
\begin{equation}
\text{Addition:}\ \ Z_{\mathcal{GP}} = Z_{\mathcal{G}} + \alpha \cdot Z_{\mathcal{P}};\quad
\text{Concate:}\ \ Z_{\mathcal{GP}} = [Z_{\mathcal{G}}; \alpha\cdot Z_{\mathcal{P}}].
\label{eqn:z_joint_simple}
\end{equation}

However, both \textit{Addition} and \textit{Concatenation} in Eqn.~(\ref{eqn:z_joint_simple}) suffer from fundamental limitations. First, neither yields a compact representation. The Zernike bases used for geometry encoding (Eqn.~\ref{eqn:shape_geometry_encode}) and pose encoding (Eqn.~\ref{eqn:pose_encode_vec}) are inherently mismatched: geometry encoding spans multiple angular frequency bands, whereas pose encoding is restricted to a single designated angular frequency. This basis mismatch leads to inefficient use of the encoding space. Second, elementwise \textit{Addition} inevitably entangles geometry and pose, causing mutual interference between the two components (as analyzed in Sec.~\ref{sec:geo_pose_discrimin}). In contrast, \textit{Concatenation} preserves disentanglement but imposes a strict encoding-length constraint: to maintain a fixed-length representation, the available dimensional budget must be split between $Z_{\mathcal{G}}$ and $Z_{\mathcal{P}}$, resulting in reduced expressiveness for both. These limitations motivate us to design a more principled and compact joint encoding strategy.

In \emph{XShapeEnc}, we propose to jointly encode shape geometry and shape pose within a shared Zernike basis, rather than combining them post hoc in coefficient space. Our key insight is that geometry and pose exhibit fundamentally different spectral behaviors: geometry energy is broadly distributed across Zernike bases, while pose energy is concentrated within a small set of angular frequency bands. To reconcile this mismatch, we embed pose as a phase modulation of geometry coefficients, which preserves geometric interpretability and avoids scale interference.

To ensure that shape pose admits non-zero projection onto the Zernike bases, we explicitly construct a harmonic pose field for each angular frequency band $m_p\in M$~(Eqn.~\ref{eqn:shape_geometry_encode}). By superposing these harmonic pose fields and combining them with the geometry mask, we form a composite geometry–pose field,

\begin{equation}
f_{\mathcal{GP}}(r,\theta) =
\bigl[f_{\mathcal{G}}(r,\theta),
\ \sum_{m_p\in M} f_{\mathcal{P}}(r,\theta; m_p, \mathbf{p})\bigr],
\label{eqn:composite_field}
\end{equation}

\noindent where $f_{\mathcal{G}}(r,\theta)$ denotes the shape geometry mask, $f_{\mathcal{P}}(r,\theta; m_p, \mathbf{p})$ denotes the harmonic pose field associated with angular band $m_p$. Projecting the geometry mask onto the Zernike bases yields complex-valued geometry coefficients, while projecting the composite pose field yields real-valued pose coefficients for the same bases. Consequently, each Zernike basis $V_n^m$ is associated with a pair of coefficients,

\begin{equation}
Z_{\mathcal{GP}} = \bigl[ (z_0^0, a_0^0), \ldots, (z_n^m, a_n^m), \ldots, (z_N^M, a_N^M) \bigr], \qquad
z_n^m \in \mathbb{C},\ a_n^m \in \mathbb{R},
\label{eqn:joint_encode_raw}
\end{equation}

A critical challenge arises from their unmatched spectral scales: geometry coefficients are typically small due to energy dispersion across many bases, whereas pose coefficients are large because harmonic pose fields concentrate energy within specific angular bands. Direct additive fusion would therefore cause pose to overwhelm geometry. Instead, we interpret the pose coefficient as a rotation angle acting on the complex geometry coefficient and perform joint encoding via phase modulation,

\begin{equation}
Z_{\mathcal{GP}} = \bigl[z_0^0\cdot e^{-i a_0^0}, \ldots, z_n^m\cdot e^{-i a_n^m}, \ldots, z_N^M\cdot e^{-i a_N^M}\bigr].
\label{eqn:joint_encode_rotate}
\end{equation}

This formulation aligns the joint encoding to the geometry coefficient scale while embedding pose information purely in the phase. Importantly, this operation preserves the coefficient magnitude and is consistent with the rotational action of Zernike bases, yielding a geometrically interpretable and invertible fusion.

\begin{wrapfigure}{r}{0.45\textwidth}
  \vspace{-5mm}
  \begin{center}
        \includegraphics[width=0.45\textwidth]{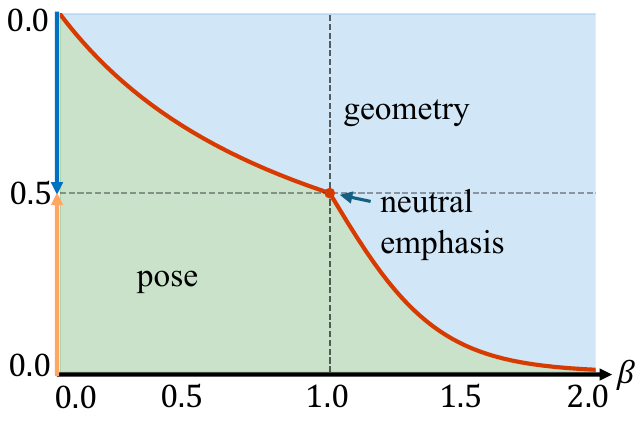}
  \end{center}
  \vspace{-4mm}
  \caption{Relative geometry-pose emphasis joint encoding visualization.}
  \label{fig:joint_encode_var}
  \vspace{-3mm}
\end{wrapfigure}

\begin{figure}[t]
    \centering
    \includegraphics[width=0.98\textwidth]{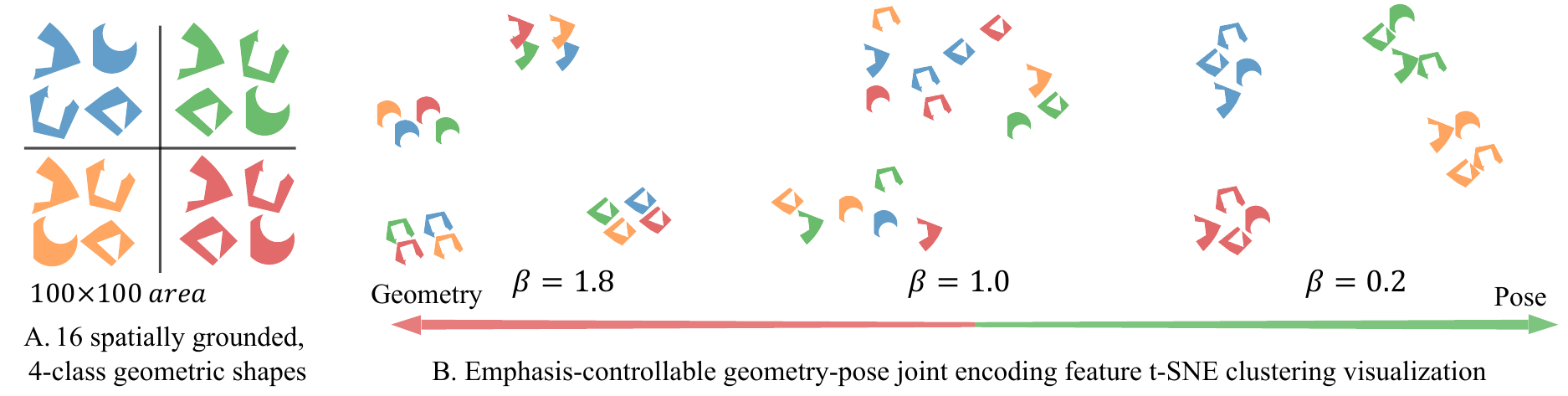}
    \caption{\emph{XShapeEnc} controllable shape geometry-pose joint encoding visualization. By varying the emphasis modulation parameter $\beta$, we can emphasize either shape geometry or shape pose.}
    \label{fig:control_geo_pose}
    \vspace{-3mm}
\end{figure}

Beyond the default joint encoding, we introduce a tunable emphasis mechanism that allows controlled bias toward either geometry or pose. We define a relative emphasis parameter $\beta \in [0, 2]$, where $\beta = 1$ indicates neutral emphasis, $\beta < 1$ emphasizes pose, and $\beta > 1$ emphasizes geometry. The more distant of $\beta$ to 1, the stronger emphasis it lays to. Intuitively, emphasizing pose corresponds to degenerating geometry coefficients toward a unit complex carrier, such that pose-induced rotations dominate. Conversely, emphasizing geometry suppresses pose-induced rotations. This yields the following formulation,

\begin{equation}
Z_{\mathcal{GP}}
=
\begin{cases}
\bigl[
e^{\beta\cdot \ln(z_0^0)}\cdot e^{-i\cdot a_0^0}, \ldots,
e^{\beta\cdot \ln(z_n^m)}\cdot e^{-i\cdot a_n^m}, \ldots,
e^{\beta\cdot \ln(z_N^M)}\cdot e^{-i\cdot a_N^M}
\bigr], & 0 < \beta \le 1,~(\text{emp. pose}) \\
\bigl[
z_0^0\cdot e^{-i \cdot \eta(\beta) a_0^0}, \ldots,
z_n^m\cdot e^{-i \cdot \eta(\beta) a_n^m}, \ldots,
z_N^M\cdot e^{-i \cdot \eta(\beta) a_N^M}
\bigr], & 1 < \beta \le 2,~(\text{emp. geometry})
\end{cases}
\label{eqn:joint_encode_emphasis}
\end{equation}

\noindent where $\eta(\beta) = \exp{(-5\cdot(\beta-1))}$ smoothly suppresses pose influence as $\beta$ increases. $\beta \in [0,1]$ emphasizes shape pose, while $\beta \in (1,2]$ emphasizes shape geometry. We show how the relative emphasis is achieved in Fig.~\ref{fig:joint_encode_var}, from which we can observe that, ranging from $0.0$ to $2.0$, $\beta$ continuously assigns different weight to shape geometry and shape pose respectively. Another qualitative example is shown in Fig.~\ref{fig:control_geo_pose}, in which we choose 4 exemplar shape geometries from \emph{XShapeCorpus} with depth $d=2$, and place each of them at the 4 sub-areas independently. As a result, each shape geometry has been duplicated at the 4 sub-areas with different shape poses, leading to a total of 16 spatially grounded shape geometries. By varying the modulation weight $\beta$ in Eqn.~(\ref{eqn:joint_encode_emphasis}), we can obtain the jointly encoding with different emphasis between shape geometry and shape pose. We test the discriminability by running t-SNE~\cite{tsne} clustering on the joint encoding under different modulation weight $\beta$ to test if the joint encoding reflects the emphasis controlled by $\beta$. We can clearly see from Fig.~\ref{fig:control_geo_pose} that we can successfully emphasize the shape geometry and shape pose by controlling $\beta$. While $\beta=1$ strikes neutral emphasis where each geometric shape is individually separated, smaller $\beta = 0.2$ and larger $\beta = 1.8$ cluster them by shape pose and shape geometry correspondingly

\textbf{Invertibility.} The joint encoding in Eqn.~(\ref{eqn:joint_encode_emphasis}) embeds shape geometry and shape pose within each complex coefficient. As a result, neither the shape geometry encoding nor the shape pose encoding can be recovered from the joint encoding alone. Nevertheless, invertibility is preserved as long as either the shape geometry encoding or the shape pose encoding is stored separately.

\subsection{Encoding Recapitulation}
\label{sec:enc_recap}

\begin{table*}[h]
\centering
\begin{minipage}[t]{0.67\textwidth}
\begin{algorithm}[H]
\caption{\emph{XShapeEnc} Algorithmic Workflow}
\label{alg:xshapeenc}
\SetKwInOut{Input}{Input}
\SetKwInOut{Output}{Output}
\Input{geometric shape $\mathcal{S}$; \textit{FreqProp} coeff. $\lambda$; encoding len. $L$; modulation weight $\beta$; optional angular freq. band $m_p$; Zernike basis $\{V_m^n\}$~(or $\{V_{m_p}^n\}$)}
decompose $\mathcal{S}$, $\mathcal{S}\rightarrow (S_{\mathcal{G}}, S_{\mathcal{P}})$, $S_{\mathcal{G}} \coloneqq f_{\mathcal{G}}$, $S_{\mathcal{P}} \coloneqq f_{\mathcal{P}}$

\If{\textit{encode shape geometry}}{
    get geometry encoding $Z_\mathcal{G} \leftarrow \mathbfcal{F}_{\mathcal{G}}(S_{\mathcal{G}})$~(Eqn.~\ref{eqn:shape_geometry_encode})\;
    optionally run \textit{FreqProp} with $\lambda$ on $Z_{\mathcal{G}}$~(Eqn.~\ref{eqn:arfreqprop})\;
    $Z_{\mathcal{G}} \leftarrow \text{complex2real}(Z_{\mathcal{G}})$~(Sec.~\ref{sec:enc_recap})\; 
    \Return $Z_{\mathcal{G}}\in \mathbb{R}^L$\;
}
\If{\textit{encode shape pose}}{
    get harmonic pose field at $m_p$, $f_{\mathcal{P}}(m_p)$~(Eqn.~\ref{eqn:harmonic_pose_field})\;
    get shape pose encoding $Z_\mathcal{P}\leftarrow \mathbfcal{F}_{\mathcal{P}}(S_{\mathcal{P}})$~(Eqn.~\ref{eqn:harmonic_field_coeff})\;
    \Return $Z_{\mathcal{P}}\in \mathbb{R}^L$\;
}
\If{\textit{joint encode geometry and pose}}{
    get harmonic pose fields $\sum_{m_p\in M} f_{\mathcal{P}}(m_p)$~(Eqn.~\ref{eqn:harmonic_pose_field})\;
    composite field $f_{\mathcal{GP}}=f_{\mathcal{G}}+\beta \sum_{m_p\in M}f_{\mathcal{P}}(m_p)$~(Eqn.~\ref{eqn:composite_field})\;
    joint encoding $Z_\mathcal{GP} \leftarrow \mathbfcal{F}_{\mathcal{GP}}(f_{\mathcal{GP}})$~(Eqn.~\ref{eqn:harmonic_field_coeff})\;
    optionally run \textit{FreqProp} with $\lambda$ on $Z_{\mathcal{GP}}$~(Eqn.~\ref{eqn:arfreqprop})\;
    $Z_{\mathcal{GP}} \leftarrow \text{complex2real}(Z_{\mathcal{GP}})$~(Sec.~\ref{sec:enc_recap})\;
    \Return $Z_{\mathcal{GP}}\in \mathbb{R}^L$\;
}
\Output{real-valued encoding: $Z_\mathcal{G}$ or $Z_\mathcal{P}$ or $Z_{\mathcal{GP}}$}
\end{algorithm}
\end{minipage}%
\hfill
% ---------- TOP TABLE ----------
\begin{minipage}[t]{0.30\textwidth}
\centering
\small
\captionof{table}{Relation between real-valued encoding length and required Zernike basis number in shape geometry Encoding and shape geometry and shape pose joint encoding~(see Sec.~\ref{sec:shape_geometry_encode} and~\ref{sec:joint_encode}).}
\label{tab:rel_len_basis}
\vspace{-2.3mm}
\begin{tabular}{c|c}
\hline
Enc. Length & Basis Num. \\
\hline
256   & 22  \\
512   & 31  \\
1024  & 44 \\
2048  & 63 \\
4096  & 91 \\
\hline
\end{tabular}

\vspace{0.5mm}  % <-- vertical gap between the two tables

\captionof{table}{Relation between real-valued encoding length and required Zernike basis number in shape pose encoding~(see Sec.~\ref{sec:shape_pose_enc}).}
\label{tab:encoding_complexity}
\begin{tabular}{c|c}
\hline
Enc. Length & Basis Num. \\
\hline
256   & 256  \\
512   & 512  \\
1024  & 1024 \\
2048  & 4096 \\
4096  & 4096 \\
\hline
\end{tabular}

\end{minipage}
\end{table*}

\emph{XShapeEnc} encoding workflow is shown in Algorithm~\ref{alg:xshapeenc}. We can learn that both \textit{step~2} and \textit{step~5} enable vectorized computation, and \textit{step~3} just requires a one-time sweep along the initial shape geometry encoding, so the whole workflow is extremely efficient with almost linear time complexity. Moreover, the Zernike basis $\{V_m^n\}$ is pre-constructed. Together with the pre-defined frequency propagation coefficient $\lambda$ and angular frequency $m_p$, the Zernike basis $\{V_m^n\}$ is used to encode arbitrary shapes.

The shape geometry encoding $Z_\mathcal{G}$ in Eqn.~(\ref{eqn:shape_geometry_encode}) is complex-valued; it serves as the foundational encoding that is fully invertible and frequency-rich. On top of $Z_\mathcal{G}$, we further extract real-valued encoding for practical needs. For example, if the goal is to obtain rotation-invariant encoding, we can simply take magnitude $|Z_\mathcal{G}|$. For rotation-variant encoding, we can adopt any pre-defined rule to flatten the $Z_\mathcal{G}$ encoding into real-valued vector encoding as long as the original $Z_\mathcal{G}$ can be recovered by reversing the rule. In order to preserve as much information as possible in Eqn.~(\ref{eqn:shape_geometry_encode}) in the final real-valued encoding, we exploit the conjugate symmetry \(z_n^{-m}=\overline{z_n^{m}}\) and retain only non-negative angular orders. Moreover, we flatten each complex-valued coefficient by putting its real and imaginary parts in juxtaposition. The relation between the final real-valued encoding length and the required Zernike basis number is given in Table~\ref{tab:rel_len_basis}, from which we can learn that we can obtain the commonly used encoding length with limited Zernike bases. For the shape pose encoding $Z_{\mathcal{P}}$ in Eqn.~(\ref{eqn:pose_encode_vec}), no complex-to-real value conversion is needed as it is already real-valued. To obtain the same encoding length as shape geometry encoding, it essentially requires twice the number of Zernike bases required by shape geometry encoding $Z_{\mathcal{G}}$.

%% file: secs_tp/sec4_exp.tex
\section{Experiment}

We exhaustively evaluate \emph{XShapeEnc} from 4 aspects:
\vspace{-3mm}
\begin{enumerate}[leftmargin=12pt]
    \item \textbf{Theoretical Validity:}  if the desired encoding properties such as invertibility, interpretability and generality are theoretically guaranteed?
    \item \textbf{Efficiency:} if the whole encoding process is computationally efficient?
    \item \textbf{Discriminability:} if \emph{XShapeEnc} encoding is discriminative for spatially grounded geometric shape representation?
    \item \textbf{Applicability:} if \emph{XShapeEnc} exhibits wide applicability in various downstream tasks?
\end{enumerate}

\subsection{XShapeCorpus Benchmark}

Currently there is no public dataset in which each 2D shape is an arbitrary geometric shape~(aka shape geometry) and further paired with a spatial position~(aka shape pose). Existing datasets such as ElementaryCQT~\cite{elementaryCQT}, AutoGeo~\cite{autogeo} and Mendeley 2D shape dataset~\cite{shape_2d_dataset} contain just simple geometric shapes~(\textit{e.g.}, triangle, square, circle) and have been associated with no spatial pose information. We hereby curate \textbf{XShapeCorpus} - a novel 2D geometric shape corpus highlighting shape diversity and pose diversity. Specifically, we depend on 8 shape primitives that are commonly seen in our daily lives:

\textit{circle}, \textit{square}, \textit{rectangle}, \textit{triangle}, \textit{diamond}, \textit{ellipse}, \textit{pentagon}, \textit{sector}

Each of the 8 shape primitives is normalized to lie within the unit disk. More complex 2D geometric shapes can be constructed by iteratively running either unary or binary shape operators. The unary shape operator operates on a single shape and contains 3 operations: \textit{scale}, \textit{translate}, \textit{rotate}. The binary shape operator operates on two shapes to create a new shape, and we incorporate 5 binary shape operators: \textit{subtract}, \textit{union}, \textit{intersect}, \textit{symmetric difference}~(\textit{xor}), \textit{convex hull}.

\begin{wrapfigure}{r}{0.45\textwidth}
\vspace{-8mm}
  \begin{center}
    \includegraphics[width=0.45\textwidth]{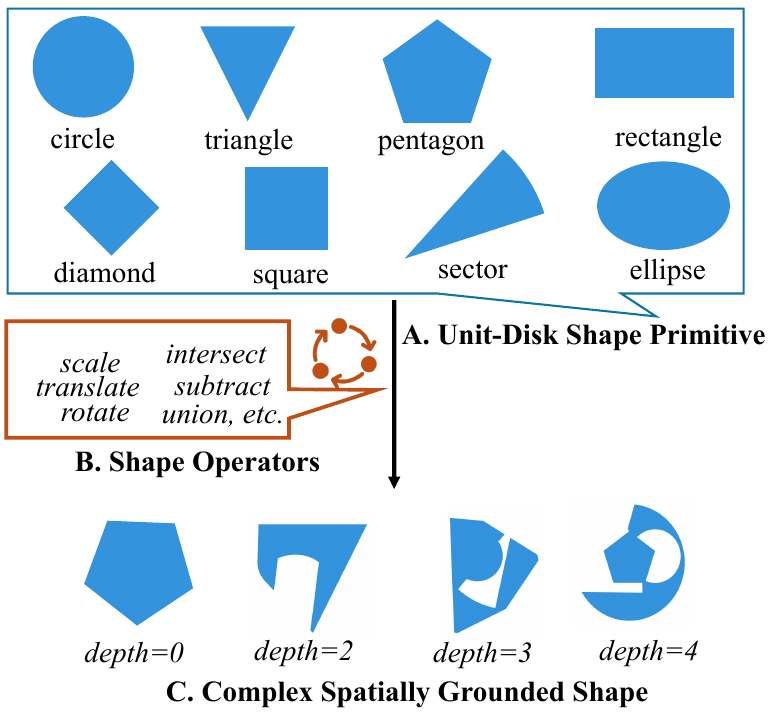}
  \end{center}
  \vspace{-4mm}
  \caption{\emph{XShapeCorpus} curation visualization. More complex shapes can be created by consecutively running shape operations. Higher depth~(operation number) indicates higher Shape complexity. Each shape is independently associated with a spatial pose.}
  \vspace{-7mm}
  \label{fig:shape_primitive}
\end{wrapfigure}

The corpus curation pipeline is shown in Fig.~\ref{fig:shape_primitive}, an intermediate shape created by the preceding operation is fed to the next randomly chosen operation to generate the new shape. As more operations generally result in higher shape complexity, we use the ``depth'', which indicates the number of operations used to construct a shape, to measure a shape's complexity. To address the potential collapse back to plain shape primitive or the existence of primitive-shaped ``tiny hole'' of the created shape, we avoid directly using a raw shape primitive during intermediate shape creation steps but instead  wrap it to match the intermediate shape fraction. Moreover, we further enforce the number of binary operations to be executed to be proportional to ``depth'' to guarantee higher depth values are mostly associated with more complex shapes. In practice, we curate \emph{XShapeCorpus} for depth ranging from 1 to 10. For each depth, we construct 100 shapes, resulting in a total of 1,000 shapes. 

\subsection{Comparing Baselines}

We exhaustively compare \emph{XShapeEnc} with 11 baselines, which can be categorized into three main classes: boundary point based encoding, regular sampled point based encoding and neural network based encoding.

For boundary point based encoding, we incorporate 4 baselines:

\begin{enumerate}
    \item \textbf{AngularSweep}. The angular sweep encoding strategy has been initially proposed and adopted by SoundTRC~\cite{soundTRC} and ReZero~\cite{rezero} to encode a 2D convex geometric shape. The underlying idea is to shoot a line from the origin and sweep counterclockwise~(or clockwise) at a predefined angular interval. The final shape geometry encoding is obtained by sequentially appending the distance value between the origin and shape boundary at each sweep angle. The original angular sweep in SoundTRC~\cite{soundTRC} and ReZero~\cite{rezero} can only encode convex shape geometry~(they assume the shooting line just intersects with the shape geometry once). We extend it to accommodate arbitrary shape geometry encoding by appending all distance values in close-to-far order at one angular interval.
    \item \textbf{2DPE}. Given shape geometry represented by a binary mask, we discretize the mask into finite grid points and adopt sinusoidal positional encoding~\cite{att_all_need} to encode each point position. The final shape geometry encoding can be obtained by mean-pooling all point encodings. \textbf{2DPE} can be used to encode both shape geometry and shape pose.
    \item \textbf{ShapeDist}~\cite{shape_distribution}, Shape distributions provide a classical training-free approach for characterizing object geometry by analyzing statistical relationships between sampled points on the object’s surface. In particular, the D2 distribution computes distances between randomly selected point pairs to form a global signature of shape. To adapt this concept to 2D silhouettes in our setting, we detect the boundary of the binary shape mask and deterministically sample boundary points at a uniform stride. We then compute all pairwise Euclidean distances among the sampled points, normalize them to ensure scale invariance, and convert them into a fixed-length histogram. The resulting descriptor acts as a compact and permutation-invariant shape representation that captures global boundary geometry without requiring learning or specialized basis functions. This serves as a strong classical baseline complementary to our proposed XShapeEnc framework.
    \item \textbf{ShapeContexts}~\cite{shape_context}. Serving as a classic 2D shape description, Shape Context~\cite{shape_context} characterizes a shape by sampling contour points and recording the relative spatial distribution of other points in a log-polar histogram around each reference point. It captures local and global geometric structure, offering robustness to moderate deformation and making it a strong baseline for shape description.
\end{enumerate}

For boundary line based encoding, we incorporate one baseline,

\begin{enumerate}
  \item \textbf{Poly2Vec}~\cite{siampou2024poly2vec}, a recent method for encoding spatially grounded polygon shapes. To extend it to arbitrary shapes, we intentionally approximate non-linear shape boundaries using sets of line segments.
\end{enumerate}

\begin{figure}[t]
    \centering
    \includegraphics[width=0.97\linewidth]{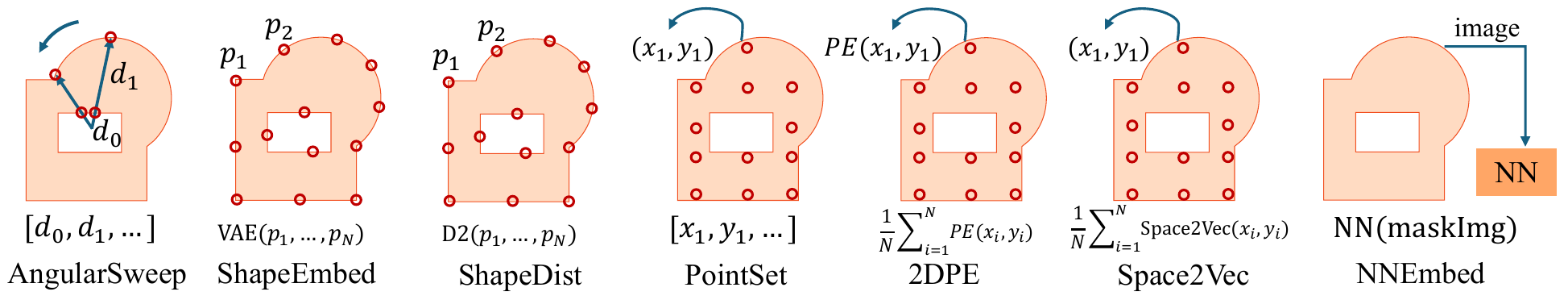}
    \caption{Typical baselines illustration: \emph{AngularSweep}, ShapeEmbed~\cite{shapeembed}, ShapeDist~\cite{shape_distribution} are based on shape geometry boundary points, the other three baselines~(PointSet, 2DPE and Space2Vec~\cite{space2vec_iclr2020}) are based on regularly sampled points.}
    \label{fig:baseline_vis}
\end{figure}

For regular-sampled point based encoding, we incorporate 3 baselines,

\begin{enumerate}
    \item \textbf{PointSet}. We approximate the geometric shape by point set by grid-sampling a set of 2D points on the shape mask. Flattening each point's $[x,y]$ coordinates and further concatenating all points' coordinates together in pre-defined order~(\textit{e.g.}, scanning each row from top to the bottom) gives us the geometric shape encoding. Point set is a well-established and flexible approach in graphics, shape analysis. In our experiment, we adopt it to represent the shape geometry.
    \item \textbf{ShapeEmbed}. ShapeEmbed~\cite{shapeembed} first converts the shape mask boundary into a translation-, rotation-, and scale-invariant Euclidean distance matrix, then trains a variational autoencoder~(VAE) to encode the matrix into latent space representation. We adopt the ShapeEmbed~\cite{shapeembed} to encode the shape geometry.
    \item \textbf{Space2Vec}. Space2Vec~\cite{space2vec_iclr2020} encodes spatial locations into fixed-dimensional vectors using a set of multi-scale sinusoidal basis functions inspired by biological grid cell responses. We incorporate it as a baseline to encode both shape geometry and shape pose. Given a point $(x,y)$ sample, Space2Vec projects it onto multiple spatial frequencies and directions to generate the encoding. To adapt Space2Vec to our setting, we approximate the shape mask by point set and compute per-point Space2Vec encoding. The final shape geometry encoding is obtained by mean average pooling all point encodings. Unlike the original Space2Vec, which is designed for general geospatial representation with optional neural projection layers, our adaptation removes all learnable components and uses a simple point sampling strategy with mean pooling, yielding a lightweight geometry representation suitable as a competitive training-free baseline for \emph{XShapeEnc}.
\end{enumerate}

For neural network based encoding~(\emph{NNEmbed}), we test 3 image-based neural networks,

\begin{enumerate}
    \item \textbf{ResNet18}~\cite{resnet18}, which is a widely used convolutional neural network model pretrained on ImageNet dataset~\cite{imagenet_dataset}.
    \item \textbf{ViT}~\cite{dosovitskiy2020vit}, which is Transformer~\cite{att_all_need} based neural network model pretrained on ImageNet dataset~\cite{imagenet_dataset}.
    \item \textbf{CLIP}~\cite{clip}, which is a neural network model pre-trained on massive aligned text-image paired data. It lays emphasis on the input image's semantics when encoding a shape. 
\end{enumerate}

We illustrate the encoding procedures underlying a subset of the baselines in Fig.~\ref{fig:baseline_vis}.

\subsection{XShapeEnc Encoding Theoretical Validity}

\begin{wrapfigure}{r}{0.4\textwidth}
\vspace{-4mm}
  \begin{center}
    \includegraphics[width=0.4\textwidth]{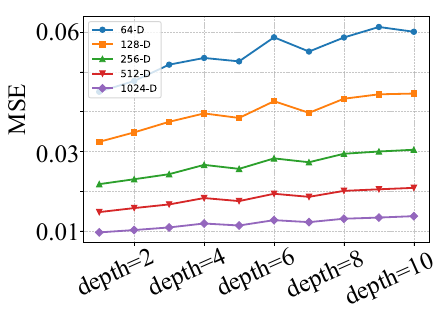}
  \end{center}
  \vspace{-4mm}
  \caption{MSE variation for encoding length over depth shape complexity on \emph{XShapeCorpus} dataset.}
  \vspace{-4mm}
  \label{fig:mse_var}
\end{wrapfigure}

\emph{XShapeEnc} is mathematically rigorous. For \emph{shape geometry}, we project the unit disk mask $f(r,\theta)$ onto the orthogonal Zernike basis $V_n^{m}(r,\theta)$ (Sec.~\ref{sec:shape_geometry_encode}), in which the orthogonality and linearity are proved in Sec.~\ref{app:sec:zernike_basis_ortho} and Sec.~\ref{sec:app:zernike_linearity_proof} in Appendix. Hence, superposition in the image domain translates exactly to superposition in coefficient space, ensuring stable and interpretable encodings. Moreover, rotations act by a phase on coefficients, giving \textbf{rotation equivariance}; taking magnitudes $|z_n^{m}|$ yields \textbf{rotation invariance} when desired (Sec.~\ref{sec:shape_geometry_encode}). To enrich spectra while preserving these properties, the \emph{frequency propagation}~(FreqProp) update in Eqn.~(\ref{eqn:arfreqprop}) is linear and \emph{invertible}. For \emph{shape pose} encoding, we specifically construct a harmonic pose field in Eqn.~(\ref{eqn:harmonic_pose_field}), and its projection onto Zernike basis is proved in Sec.~\ref{sec:app:harmonic_posefield_derive} in Appendix. With radially orthonormal windows $\{w_k\}$ and $K \leq L$, the induced matrix $C$ is full-rank and well-conditioned, guaranteeing \emph{invertible} recovery of $\mathbf{p}$ from the shape pose encoding. Therefore, the whole encoding is fully valid and mathematically grounded.

\begin{figure}[h]
    \centering
    \includegraphics[width=0.98\linewidth]{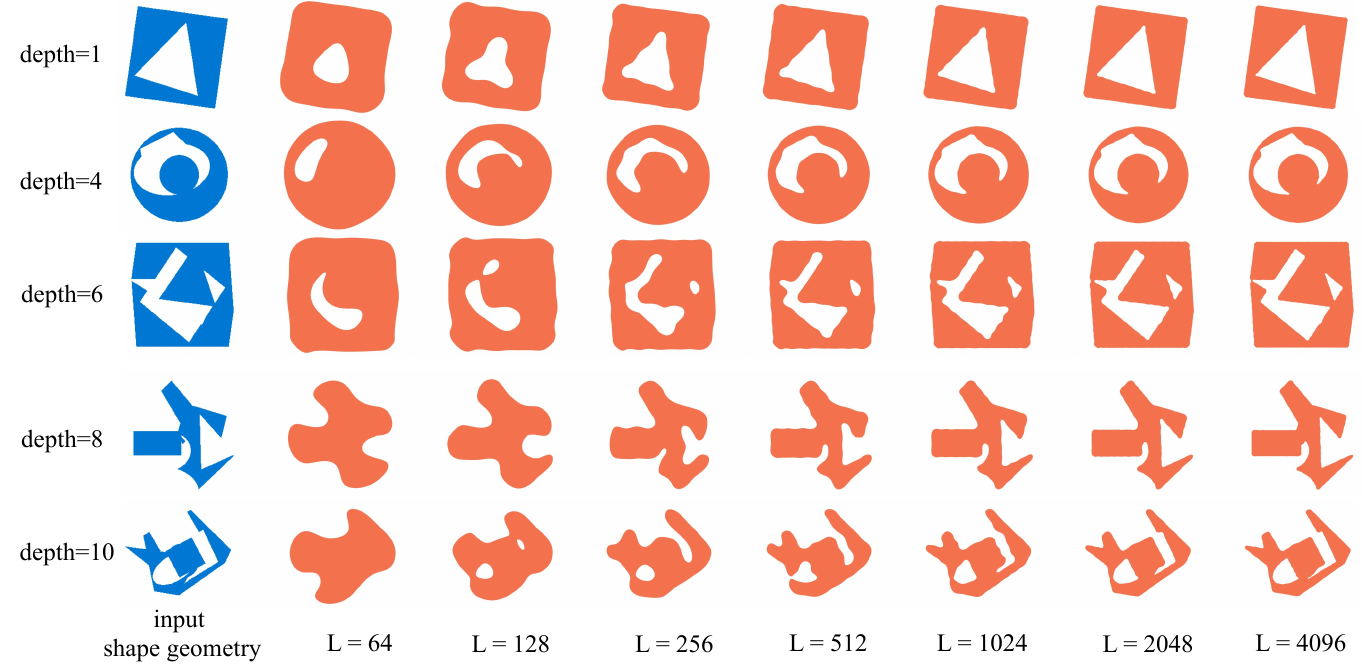}
    \caption{\emph{XShapeEnc} encoding invertibility illustration. We show two complex shape geometry~(within unit disk) with depth = 5 and depth = 10 and their reconstructed shape geometry under various encoding length. Note that the original reconstructed shape is soft-masked, we binarize it with the threshold $0.2$ for better visualization.}
    \label{fig:invert_vis}
\end{figure}

\textbf{Shape Geometry Encoding Invertibility}. Given the constructed shape corpus \emph{XShapeCorpus}, we exhaustively test shape geometry reconstruction error~(mean square error~MSE) under various encoding lengths and shape geometry complexity. To this end, first, we encode each shape geometry by setting the rasterization resolution as 300, and use target encoding lengths covering 64, 128, 256, 512, 1024, 2048, 4096. After applying the inversion process, we compute the shape geometry reconstruction error under various target lengths and shape complexity informed by depth. Five qualitative shape geometry reconstructions under the 7 encoding lengths on 5 complex shapes~(depth=$1,4,6,8,10$) are shown in Fig.~\ref{fig:invert_vis}, from which we can learn that the input complex geometric shape can be reconstructed from the shape geometry encoding. The larger the encoding length, the higher the reconstruction accuracy. The quantitative reconstruction MSE variation w.r.t. different shape complexity~(indicated by depth) under various encoding length is shown in Fig.~\ref{fig:mse_var}. We observe from this figure that longer encoding length results in smaller MSE values, while more complex shapes lead to higher MSE than simpler ones.  This observation is consistent with \emph{XShapeEnc} encoding principle where the input shape geometry is essentially approximated by a set of orthogonal Zernike basis. More complex shape geometry requires more bases, and encoding length cutoff inevitably leads to higher reconstruction error.

\textbf{Shape Pose Encoding Invertibility}. We find the shape pose can be precisely reconstructed from any of the encoding length we test for shape geometry encoding~(the MSE=0).  In fact, based on the theoretical analysis in Sec.~\ref{eqn:pose_encode_matrix} and Eqn.~(\ref{eqn:pose_encode_matrix}), we can mathematically derive that the pose vector can be precisely recovered  as long as the encoding length is larger than the pose parameter number.
 
\subsection{XShapeEnc Encoding Efficiency}

Based on the discussion in Sec.~\ref{sec:shape_geometry_encode} and Sec.~\ref{sec:enc_recap}, we can conclude that \emph{XShapeEnc} is extremely efficient as:

\begin{enumerate}
\item it is training-free -- no learning process is needed.
\item benefiting from the \textit{Linearity} property, we can directly composite each single geometric shape encoding together to get the complex shape encoding, without requiring to encode the complex composite shape from scratch.
\item as is shown in Algorithm~\ref{alg:xshapeenc}, the Zernike basis just needs to be constructed once, the shape geometry and shape pose encoding can be executed in parallel, within each of which vectorized computation can be leveraged to speed up the computation. The sequential one-time sweep computation is just required in frequency propagation.
\end{enumerate}

\subsection{XShapeEnc Encoding Discriminability}
\label{sec:discrimin}

\begin{figure}[h]
  \begin{center}
    \includegraphics[width=0.999\textwidth]{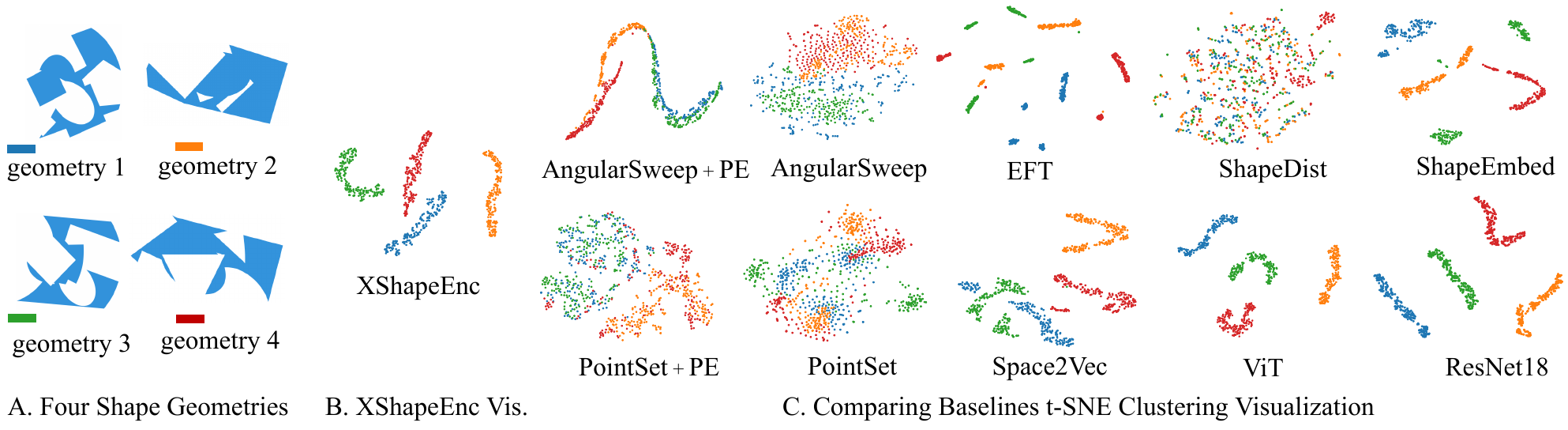}
  \end{center}
  \vspace{-4mm}
  \caption{Shape geometry t-SNE~\cite{tsne} clustering visualization between \emph{XShapeEnc} and other baselines. We choose four main complex shape geometries~(subfig.~A) with depth = 10 and augment each one to obtain 200 variations by operations including rotation, shearing and elastic deformation. We compare \emph{XShapeEnc} with both boundary-based shape representation~(AngularSweep w/o w/ positional encoding, elliptical Fourier transform~(EFT~\cite{ellip_fourier_contour}, ShapeDistribution~(ShapeDist~\cite{shape_distribution}, ShapeEmbed~\cite{shapeembed}), point-set shape representation~(PointSet w/ w/o positional encoding, Space2Vec~\cite{space2vec_iclr2020}) and ImageNet pretrained model, ResNet18~\cite{resnet18} and ViT~\cite{dosovitskiy2020vit}.}
  \vspace{-4mm}
  \label{fig:shape_tsne}
\end{figure}

\subsubsection{Shape Geometry Encoding Discriminability}
\label{sec:geo_discrimin}

To assess \emph{XShapeEnc} shape geometry encoding discriminability, we test if the encoding is capable of maintaining inter-class shape geometry separability when intra-class shape augmentation exists. To this end, we choose four complex shape geometries with depth=10 from the \emph{XShapeCorpus}. For each shape geometry, we explicitly apply rich shape augmentation methods to add shape variation to it. Typical augmentations include random rotation, shearing, vertex jitter and elastic deformation. We run shape augmentation for each shape geometry 200 times independently, resulting in a total of 400 shape geometries. By encoding each of these 800 shape geometries by both \emph{XShapeEnc} and other relevant baselines, we obtain the shape geometry encoding~(in our case, the encoding length is $512$) for the augmented four shape geometries, with which we further run t-SNE~\cite{tsne} for each method to cluster those encodings to test if the encodings still maintain inter-shape separability and intra-shape cohesion when intra-shape geometry perturbation presents. The clustering result is shown in Fig.~\ref{fig:shape_tsne}.

Each mask within the unit-disk of each shape is treated as a shape geometry and further gets encoded to obtain a 512~$d$ feature. We take the magnitude to obtain rotation-invariant feature, and further run t-SNE~\cite{tsne} to cluster these features. As is shown in Fig.~\ref{fig:shape_tsne} A, the well separated clusters indicate strong inter-class separability, while clusters indicate intra-class cohesion despite the augmentation. For comparison, we compare with another training-free discrete 2D positional encoding~(2D PE)~\cite{att_all_need}, in which we first discretize each augmented shape geometry mask into $300\times 300$ points, and then use the sinusoidal positional encoding to encode each shape mask~(indicated by 1) $x$ coordinate and $y$ coordinate into 256~$d$ feature before concatenating them together to form 512~$d$ feature. The clustering is shown in Fig.~\ref{fig:shape_tsne} B, we can clearly observe that \textit{2D PE} loses the inter-class shape geometry separability capability and mixes all shapes together.

\begin{wrapfigure}{r}{0.4\textwidth}
  \begin{center}
  \vspace{-5mm}
    \includegraphics[width=0.4\textwidth]{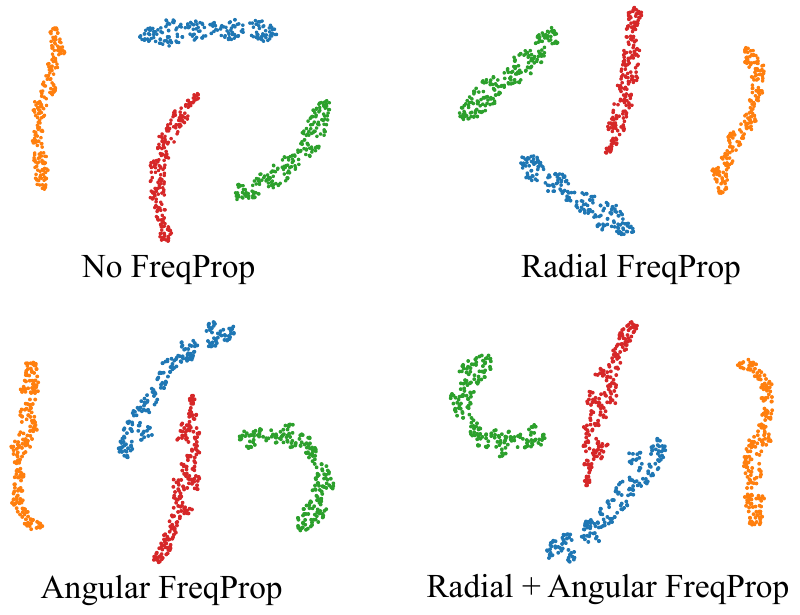}
  \end{center}
  \vspace{-5mm}
  \caption{Shape geometry encoding comparison w/ w/o FreqProp.}
  \label{fig:freqprop_compare}
\end{wrapfigure}

To analyze the effect of FreqProp on shape geometry discriminability, we further compare t-SNE clustering results with and without frequency propagation. As shown in Fig.~\ref{fig:freqprop_compare}, both radial FreqProp and angular FreqProp preserve clear inter-class separation while maintaining compact intra-class structure, indicating that propagation does not distort the underlying geometric identity captured by the encoding. More importantly, FreqProp improves frequency diversity without sacrificing class-level organization in the latent space, suggesting a favorable trade-off between representational richness and discriminative stability. This behavior is consistent with our formulation: propagation redistributes information across neighboring Zernike modes while retaining structured harmonic relationships. In other words, Zernike basis encoding provides a stable geometric backbone, and FreqProp acts as a controlled enhancement mechanism that strengthens downstream learnability while preserving shape-aware discriminability.

\begin{figure}[t]
    \centering
    \begin{minipage}{0.44\textwidth}
        \centering
        \includegraphics[width=\linewidth]{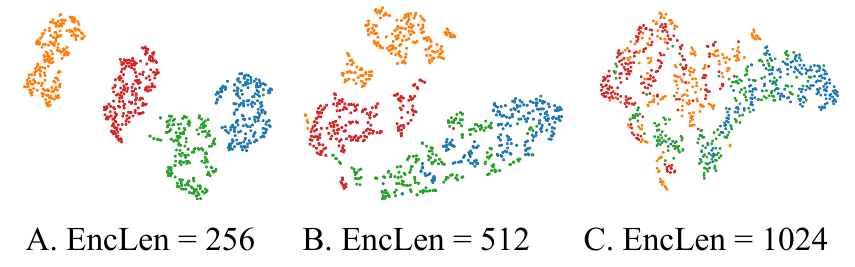}
        \caption{t-SNE~\cite{tsne} clustering result visualization on various encoding length}
        \label{fig:encode_len_impact}
    \end{minipage}\hfill
    \begin{minipage}{0.52\textwidth}
        \centering
        \includegraphics[width=\linewidth]{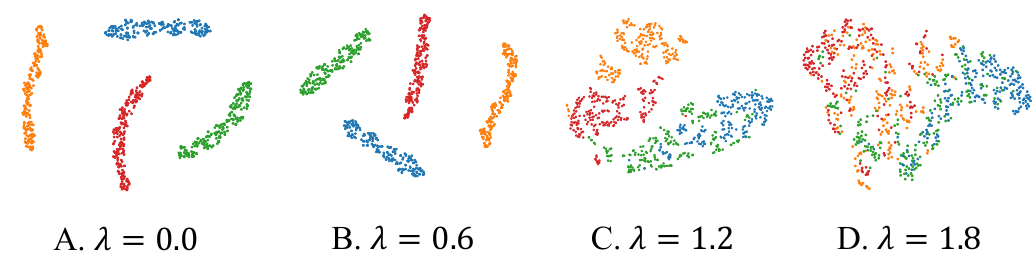}
        \caption{t-SNE~\cite{tsne} clustering result visualization on frequency propagation coefficient $\lambda$.}
        \label{fig:lambda_test}
    \end{minipage}
\end{figure}

\begin{wrapfigure}{r}{0.45\textwidth}
    \vspace{-10pt}
    \centering
    \includegraphics[width=\linewidth]{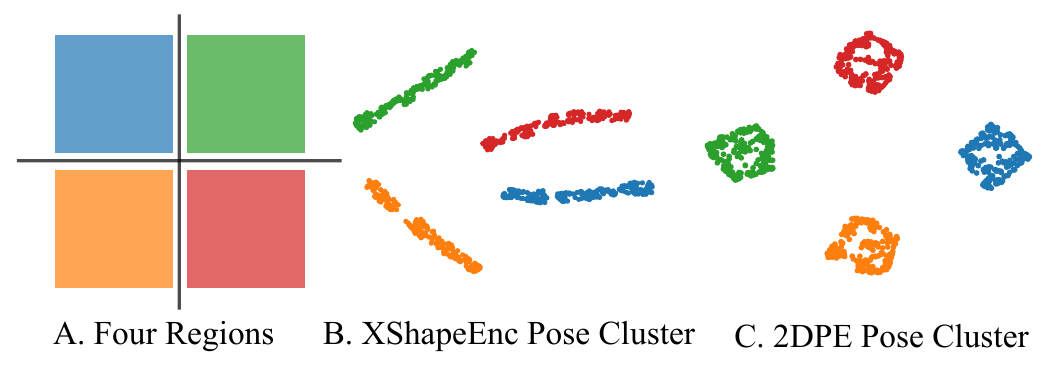}
    \caption{Shape pose t-SNE~\cite{tsne} clustering visualization. We divide a region into four main sub-regions and sample 200 points at each sub-region~(subfig.~A). Both \emph{XShapeEnc} and classic 2D positional encoding can successfully cluster shape poses based on their spatial position.}
    \label{fig:shape_pose_cluster}
    \vspace{-10pt}
\end{wrapfigure}

To further test the impact of encoding length, we further test the clustering result with encoding length in 256, 512 and 1024. From the result shown in Fig.~\ref{fig:encode_len_impact}, we can observe that, under shape augmentation, longer shape encoding length introduces more encoding feature difference. This is anticipated because longer encoding length captures more localized shape feature. Shape augmentation pronounces such localized shape feature variation. We further ablate the frequency propagation coefficient $\lambda$ on the shape geometry discriminability. By varying $\lambda \in [0.0, 0.6, 1.2, 1.8]$, we visualize the corresponding clustering result in Fig.~\ref{fig:lambda_test}. We can clearly see higher $\lambda$ introduces larger portion of low-frequency coefficient to high-frequency coefficient, the resulting encoding gradually loses shape discriminability.

\subsubsection{Shape Pose Encoding Discriminability}
\label{sec:pose_discrmin}

To assess \emph{XShapeEnc} shape pose encoding discriminability, we construct a $100\times 100$ $m^2$ area and evenly divide it into 4 sub-areas: \textit{topleft}, \textit{topright}, \textit{bottomleft} and \textit{bottomright}. Within each sub-area, we randomly sample 200 shape poses and use \emph{XShapeEnc} and 2D positional encoding~\cite{att_all_need} to encode each shape pose into 512~$d$ features. The t-SNE~\cite{tsne} clustering results for \emph{XShapeEnc} and \textit{2D PE} encoding are shown in Fig.~\ref{fig:shape_pose_cluster}. From this figure, we can learn that both \emph{XShapeEnc} and \textit{2D PE} show shape pose discriminability.  Compared to \textit{2D PE}, which mainly preserves absolute coordinate values, \emph{XShapeEnc} organizes poses along continuous low-dimensional manifolds (belt-like trajectories in t-SNE). This reveals that \emph{XShapeEnc} embeds spatial transformations in a geometrically consistent manner—small pose perturbations correspond to small latent displacements, while different pose branches remain clearly separated. In summary, \emph{XShapeEnc} exhibits strong encoding discriminability for both shape geometry and pose encoding.

\subsubsection{Shape Geometry and Pose Joint Encoding Discriminability}
\label{sec:geo_pose_discrimin}

\begin{figure}[h]
  \begin{center}
    \includegraphics[width=0.99\textwidth]{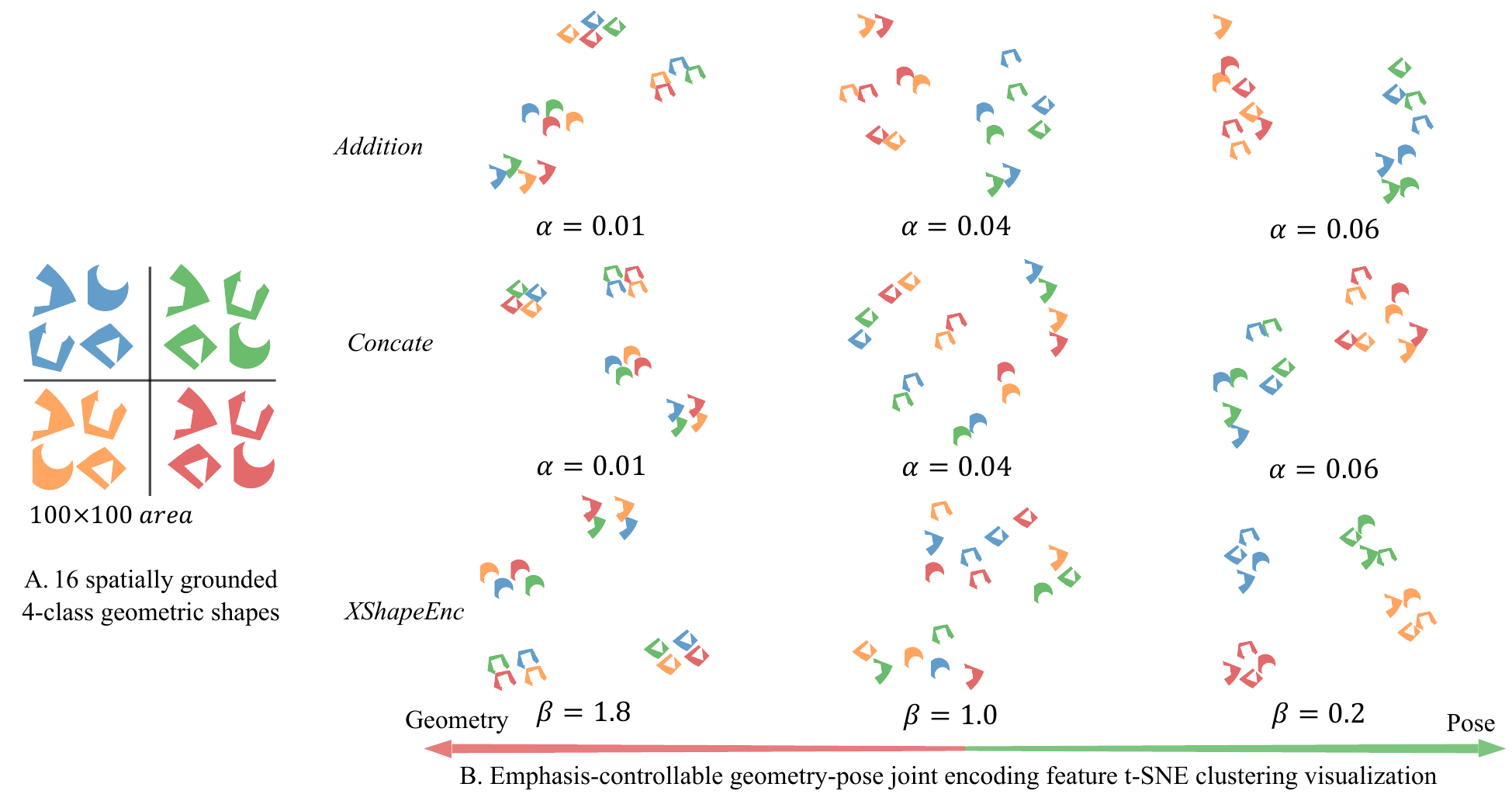}
  \end{center}
  \caption{Shape Geometry and Pose Joint Encoding tSNE~\cite{tsne} clustering result visualization}
  \label{fig:geometrypose_cluster}
\end{figure}

To assess the joint encoding discriminability, we choose 4 exemplar shape geometries from \emph{XShapeCorpus} with depth $d=2$, and further place each of the 4 shape geometries at the 4 sub-areas in Sec.~\ref{sec:pose_discrmin} independently. As a result, each shape geometry has been duplicated at the 4 sub-areas but with different shape pose~(\textit{e.g.}, $x$- and $y$- coordinates), leading to a total of 16 spatially grounded shape geometries. We then encode each of the spatially grounded shape geometry by the shape geometry and pose jointly encoding strategy as presented in Sec.~\ref{sec:joint_encode}. By varying the modulation weight $\beta$ in Eqn.~(\ref{eqn:joint_encode_emphasis}), we can obtain the jointly encoding with different emphasis between shape geometry and shape pose. We test the discriminability by running t-SNE~\cite{tsne} clustering on the joint encoding under different modulation weight $\beta$ to test if the joint encoding reflects the emphasis governed by $\beta$. Meanwhile, we compare the other two straightforward joint encoding methods presented in Eqn.~(\ref{eqn:z_joint_simple}): \textit{Addition} and \textit{Concatenation}.

The clustering result is shown in Fig.~\ref{fig:geometrypose_cluster}, from which we can clearly see that: smaller $\beta$ results in the joint encoding emphasizing on shape geometry, while larger $\beta$ emphasizes on shape pose. It thus shows the discriminability in our proposed shape geometry and shape pose joint encoding.

\subsection{XShapeEnc Encoding Applicability}

\subsubsection{Shape Retrieval}

\begin{table}[h]
    \small
    \centering
    \caption{2D shape retrieval result on Mendeley 2D shape dataset~\cite{shape_2d_dataset} and MPEG-7 CE-Shape-1 Part B dataset~\cite{latecki2006mpeg7}. We report mean Average Precision~(mAP) by aggregating average prevision~(AP) of querying each shape in the datasets. The encoding length is $512$. The top-1, top-2 and top-3 performing methods are labeled with different color.}
    \begin{tabular}{l|l|cc}
    \hline
      Description & Method Name &  Mendeley Shape & MPEG-7 Shape \\
    \hline
   \multirow{5}{*}{Training-free} & PointSet & 0.21 & 0.10\\
    & ShapeContexts~\cite{shape_context} & 0.55 & 0.45 \\
    & Space2Vec~\cite{space2vec_iclr2020} & 0.30 & 0.47 \\
    & ShapeDist~\cite{shape_distribution} & 0.25 & 0.11 \\
    & AngularSweep~\cite{soundTRC} & 0.42 & 0.26 \\
    \hline
    Training-required & ShapeEmbed~\cite{shapeembed} & \cellcolor{thirdcolor}0.73 & 0.41 \\
    \hline
    \multirow{3}{*}{Pretrained Model} &
    ResNet18~\cite{resnet18} & 0.40 & 0.56 \\
    &ViT~\cite{dosovitskiy2020vit} & 0.43 & 0.57 \\
    &CLIP~\cite{clip} &  0.57 & \cellcolor{topcolor}0.64 \\
    \hline
    \multirow{3}{*}{Ours} &
    \emph{XShapeEnc}~(FreqProp+Comp2Real) & 0.53 & 0.54 \\
    &\emph{XShapeEnc}~(FreqProp+Mag.) & \cellcolor{secondcolor}0.88 & \cellcolor{thirdcolor}0.58\\
    & \emph{XShapeEnc}~(Mag.) & \cellcolor{topcolor}\textbf{0.91} & \cellcolor{secondcolor}\textbf{0.59} \\
    \hline
    \end{tabular}
    \label{tab:shape_retrieval_result}
\end{table}

We first evaluate \emph{XShapeEnc} shape geometry encoding capability on the classic 2D shape task: shape retrieval. In shape retrieval, each shape~(in our case, shape geometry) is represented within an image. Given a query shape image, the target is retrieve images of the same class from the gallery set by computing pairwise encoded feature similarity~(in our case, we use $\cos()$ similarity). For evaluation metric, we adopt mean average precision~(mAP) which aggreates the average precision~(AP) of querying each shape image. It reflects both retrieval accuracy and ranking quality, and serves as a general and widely adopted metric for shape retrieval performance.
 
We run shape retrieval on two public datasets: Mendeley 2D shape benchmark~\cite{shape_2d_dataset} and MPEG-7 CE-Shape-1 Part B dataset~\cite{latecki2006mpeg7}. Mendeley 2D shape benchmark consists of 9 primitive shape classes: Triangle, Square, Pentagon, Hexagon, Heptagon, Octagon, Nangon, Circle and Star. Each class is instantiated with 10,000 shapes with large variations in scale, rotation and deformation. MPEG-7 CE-Shape-1 Part B dataset~\cite{latecki2006mpeg7} contains 70 organic shape classes, ranging from animal shape~(\textit{e.g.}, cattle, beettle, camel, butterfly), people~(\textit{e.g.}, children, face), utensils~(\textit{e.g.}, fork, spoon, jar) to devices~(\textit{e.g.}, watch, cellular phone). Each class associates with 20 images, resulting in a total of 1,400 images. The two datasets test the shape geometry encoding generalization capability in handling both shape geoemtry primitives and real-world shape geometries. We compare \emph{XShapeEnc} with 9 baselines, comprehensively covering training-free, training-required and pre-trained models. Within \emph{XShapeEnc}, we evaluate three variants: 1. \emph{XShapeEnc} with FreqProp~(see Sec.~\ref{sec:freqprop}) and complex-to-real conversion~(FreqProp + Comp2Real), the resulting feature is rotation-variant and thus not ideal for the shape retrieval task by design; 2. \emph{XShapeEnc} with FreqProp followed by taking the magnitude of the complex encoding~(Eqn.~\ref{eqn:arfreqprop}) as the final feature, which is approximately rotation invariant because FreqProp process relatively breaks the rotation invariance~(FreqProp + Mag.); 3. \emph{XShapeEnc} without FreqProp and taking the magnitude from the complex encoding~(Eqn.~\ref{eqn:shape_geometry_encode}) as the final feature~(Mag.).

The quantitative retrieval results are reported in Table~\ref{tab:shape_retrieval_result}. Several clear findings emerge. First, \emph{XShapeEnc}~(Mag.) achieves the best mAP ($0.91$) on the Mendeley shape dataset and the second-best mAP ($0.59$) on the MPEG-7 CE-Shape-1 Part B dataset, substantially outperforming most baselines. This indicates that \emph{XShapeEnc} captures global shape structure more consistently under large intra-class variation. Second, \emph{XShapeEnc}~(FreqProp+Mag.) also performs strongly (mAP $0.88$), suggesting that frequency enrichment remains effective even when propagation introduces slight rotation sensitivity. Third, while all non-pretrained baselines show performance drops on MPEG-7 CE-Shape-1 Part B~\cite{latecki2006mpeg7}, all three pretrained models show performance gains. We attribute this trend to the stronger correlation between MPEG-7 categories and natural-image semantics (\text{e.g.}, ImageNet~\cite{imagenet_dataset}) used during pretraining. Nevertheless, \emph{XShapeEnc} remains highly competitive across datasets with very different shape distributions.

\subsubsection{Inter-Shape Topological Relation Classification}

\begin{figure}[t]
    \centering
    \includegraphics[width=0.98\linewidth]{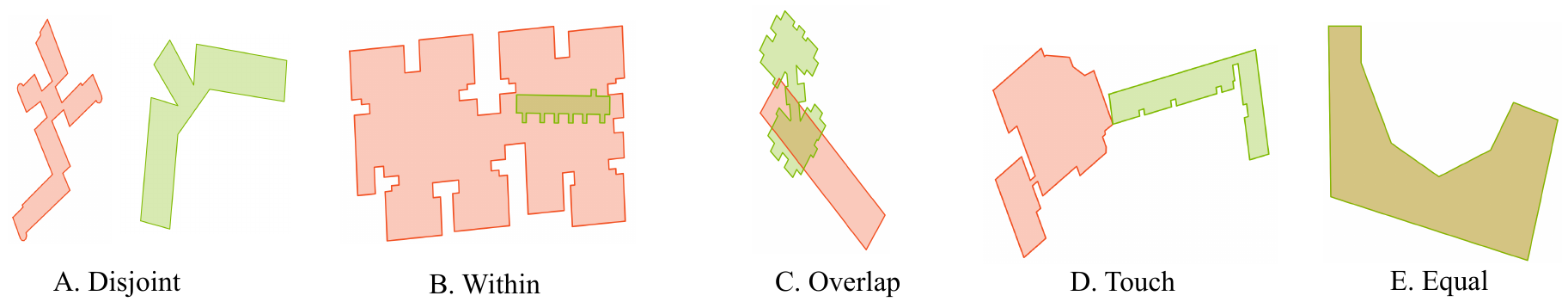}
    \caption{Polygon-polygon topological relation visualization: we visualize 5 inter-shape topological relation in OpenStreetMap Singapore dataset: \textit{Disjoint}, \textit{Within}, \textit{Overlap}, \textit{Touch} and \textit{Equal}. \textit{Equal} means two polygons are identical. From these visualizations, we can learn that these polygons vary drastically in terms of shape geometry complexity, size and scale.}
    \label{fig:top_relation_vis}
\end{figure}

To evaluate shape geometry and shape pose joint encoding capability, we follow Poly2Vec~\cite{siampou2024poly2vec} to run experiments on inter-shape topological relationship classification task~\cite{topo_relation_define}. The data is from geospatial OpenStreetMap~\cite{geospatial_big_data} in two cities: New York and Singapore, in which each shape is a 2D polygon associated with a spatial position. The 2D polygons indicate the building contour structure from the bird's-eye view~(BEV), so they vary significantly w.r.t. polygon size, shape structure and orientation. Two such polygon shapes jointly present 5 potential topological relations: \textit{disjoint}, \textit{touch}, \textit{equal}, \textit{overlap} and \textit{within}~(see Fig.~\ref{fig:top_relation_vis}). The shape geometry and shape pose of the two shapes intertwine to inform the topological relation, so the topological relation classification task serves as an ideal testbed to test the geometry-pose joint encoding capability.

Following the configuration of Poly2Vec~\cite{siampou2024poly2vec}, for each topological relation we construct 3,000 polygon-polygon pairs for training and 1,000 pairs for testing, resulting in 15,000/3,000 training/test pairs for both New York and Singapore. For all baselines except Poly2Vec~\cite{siampou2024poly2vec}, we represent the two shape masks involved in a topological relation within a single image. When the two masks are too far apart~(\textit{e.g.}, in the \textit{disjoint} case), we intentionally move them closer so that both can be enclosed within one image while preserving their topological relation. In \emph{XShapeEnc}, each spatially grounded polygon is encoded into a 512-$d$ feature with $\beta = 0.2$. For fair comparison, we follow Poly2Vec~\cite{siampou2024poly2vec} and use a two-layer multi-layer perceptron~(MLP) to predict the logits. The whole network is trained with cross-entropy loss using the Adam optimizer with a learning rate of $0.001$.

\begin{wraptable}{r}{0.47\textwidth}
    \centering
    \small
    \caption{Polygon-polygon topological relation classification accuracy~($\uparrow$).}
    \vspace{-0.5em}
    \begin{tabular}{l|cc}
        \toprule
        Method & Singapore & New York \\
        \midrule
        PointSet & 0.670 & 0.564\\
        ShapeContexts~\cite{shape_context} & 0.581 & 0.525\\
        AngularSweep~\cite{soundTRC} & 0.606 & 0.546\\
        Space2Vec~\cite{space2vec_iclr2020} & 0.706 & 0.632 \\
        \midrule
        ResNet18~\cite{resnet18} & 0.674 & 0.753\\
        ViT~\cite{dosovitskiy2020vit} & 0.669 & 0.752\\
        CLIP~\cite{clip} & 0.700 & 0.779\\
        \midrule
        Poly2Vec~\cite{siampou2024poly2vec} & 0.702 & 0.684 \\
        \midrule
        \emph{XShapeEnc}~(Ours)  & \cellcolor{topcolor} 0.760 & \cellcolor{secondcolor} 0.768 \\
        \bottomrule
    \end{tabular}
    \label{tab:topo_rel_classify}
\end{wraptable}

The classification accuracies are reported in Table~\ref{tab:topo_rel_classify}. Several observations support the effectiveness of \emph{XShapeEnc}. First, on the Singapore split, \emph{XShapeEnc} achieves the best accuracy ($0.760$), outperforming the strongest competing baseline CLIP~\cite{clip} ($0.700$) by $0.060$, as well as Poly2Vec~\cite{siampou2024poly2vec} ($0.702$) and Space2Vec~\cite{space2vec_iclr2020} ($0.706$). This margin is meaningful because the task depends not only on local contour cues but also on jointly reasoning about global geometry and relative pose. Second, on the New York split, \emph{XShapeEnc} reaches $0.768$, remaining highly competitive with the best-performing CLIP model ($0.779$) and outperforming Poly2Vec ($0.684$) by a large margin. These results suggest that \emph{XShapeEnc} provides a strong geometry--pose representation without relying on large-scale pretraining, and that its advantage is especially pronounced when topological reasoning must be inferred directly from structured spatial shape relations rather than high-level semantic priors.

\subsubsection{Spatial Target Region Control Task}

Spatial acoustic target region control task has been recently proposed in~\cite{soundTRC,rezero,create_speech_zone}, in which it aims to extract the spatial audio within a pre-specified 2D target region from the audio mixture. This task serves as an ideal scenario to test the 2D spatially grounded geometric shape encoding capability because its output is solely conditioned on the specified target region that accounts for both shape geometry and shape pose. We follow the setting used by SoundTRC~\cite{soundTRC} to focus on three kinds of spatially grounded geometric shapes: \textit{Angle}, \textit{Distance} and their combination~\textit{Angle-Distance}. By varying the angle and distance for each geometric shape, we can construct a set of indoor environment 2D geometric shapes on the floorplan plane, each geometric shape is spatially grounded to the 3D indoor environment center. While following the setting in~\cite{soundTRC}, we constrain all sound sources to lie on the same plane and adopt Pyroomacoustics~\cite{pyroomacoustics} to synthesize the data. Specifically, we have created 30-hour training audio dataset and 6-hour test audio dataset. In evaluation, we report signal-to-distortion ratio (SDR)~\cite{sdr_eval}, short-time objective intelligibility (STOI)~\cite{stoi_eval}, perceptual evaluation of speech quality (PESQ)~\cite{PESQ_eval}. We compare \emph{XShapeEnc} with ShapeEmbed~\cite{shapeembed}, Poly2Vec~\cite{siampou2024poly2vec}~(the ring-like shape is approximated by polygons) and SoundTRC angular sweeping encoding~\cite{soundTRC}. 

\begin{wraptable}{r}{0.55\textwidth}
    \centering
    \small
    \caption{Quantitative result on TRC task across three shapes: \textit{Angle}, \textit{Distance} and \textit{Angle-Dist}.}
    \begin{tabular}{l|cccc}
      \toprule
      Encoding Method & SDR~($\uparrow$) & STOI~($\uparrow$) & PESQ~($\uparrow$)  \\
      \hline
      SoundTRC~\cite{soundTRC} & 14.10 & 0.82 & 2.31  \\
      Space2Vec~\cite{space2vec_iclr2020} & 15.00 & 0.83 & 3.40 \\
      Poly2Vec~\cite{siampou2024poly2vec} &  15.10 & 0.89 & 3.90 \\
      ShapeEmbed~\cite{shapeembed} & 15.37& 0.97 & 4.26\\
      \midrule
      \emph{XShapeEnc}~(Ours) & \cellcolor{topcolor}\textbf{17.62} & \cellcolor{topcolor}\textbf{0.98} & \cellcolor{topcolor} \textbf{4.38}  \\
      \bottomrule
    \end{tabular}
    \label{tab:soundtrc_quan_rst}
\end{wraptable}

For target region control, we have synthesized 30-hour and 6-hour audio training and test, respectively. The sampling frequency is 16~kHz and each audio sample is 3 seconds long. The IRM mask~(ideal ratio mask) to be learned is of shape $160\times 320$. For both ShapeEmbed~\cite{soundTRC} and XShapeEnc, we directly use SoundTRC~\cite{soundTRC} neural network. For Poly2Vec~\cite{siampou2024poly2vec}, we follow its original setting to add an MLP layer to fuse the magnitude and phase features. We train all models on 8 NVIDIA A100 GPUs and train them for 150 epochs~(afterwards, we find the performance gradually decays). Adam optimizer~\cite{kingma2015adam} is used with an initial learning rate 0.001, which decays every 50 epochs with a decay rate of 0.5.

The quantitative results are reported in Table~\ref{tab:soundtrc_quan_rst}, where \emph{XShapeEnc} consistently achieves the best performance across all three evaluation metrics. In particular, \emph{XShapeEnc} reaches an SDR of $17.62$, outperforming the strongest baseline ShapeEmbed~\cite{shapeembed} ($15.37$), as well as Poly2Vec~\cite{siampou2024poly2vec} ($15.10$), Space2Vec~\cite{space2vec_iclr2020} ($15.00$), and SoundTRC~\cite{soundTRC} ($14.10$). Similar advantages are also observed for STOI and PESQ, where \emph{XShapeEnc} achieves the top scores of $0.98$ and $4.38$, respectively. These gains are particularly meaningful because the target region control task depends on precise modeling of both shape geometry and spatial pose: even small encoding errors can degrade the alignment between the specified region and the extracted audio. The strong and consistent improvements therefore provide direct evidence that \emph{XShapeEnc} offers a more faithful and task-relevant representation of spatially grounded geometric shapes, making it highly practical for downstream tasks built upon 2D geometric shape modeling.

%% file: secs_tp/sec5_conclusion.tex
\section{Conclusion and Discussion} 

In this work, we introduced \emph{XShapeEnc}, a unified, training-free, and task-agnostic framework for encoding arbitrary 2D spatially grounded geometric shapes. Unlike existing shape-encoding approaches, which either treat shape representation as a byproduct of a task-specific learning pipeline or focus narrowly on shape recognition settings where important attributes such as size, spatial position, and scale are intentionally ignored, \emph{XShapeEnc} formulates 2D shape encoding as an independent problem and accommodates diverse practical encoding requirements. Built upon the classical Zernike basis~\cite{zernike_moment}, \emph{XShapeEnc} decomposes a spatially grounded geometric shape into normalized within-unit-disk shape geometry and a spatial pose vector, which is further expressed as a Zernike-compatible harmonic pose field. To ensure flexibility and controllability, \emph{XShapeEnc} supports separate or joint encoding of shape geometry and pose, with tunable emphasis between the two. In addition, an optional frequency-propagation step enriches high-frequency content without compromising invertibility or structural fidelity.

Looking forward, we hope this work sheds light to a different research direction other than the overwhelming data-driven and neural network centered direction: one in which strong inductive structure, explicit geometry, and theoretical transparency play a central role. \emph{XShapeEnc} suggests that non-learning, non-data-driven encoding strategies can still be competitive when they are carefully aligned with the underlying mathematical structure of the problem. We believe this insight is valuable not only for geometric shape modeling, but also for future studies of spatially grounded intelligence in vision, graphics, acoustics, robotics, and multimodal learning. In this sense, \emph{XShapeEnc} is not only a practical encoding method, but also a step toward a more general framework for representing structured 2D spatial information.

%% file: secs_tp/sec6_appendix.tex
\section{Appendix}

\subsection{Zernike Basis Orthogonality Proof}
\label{app:sec:zernike_basis_ortho}
\begin{proof}
Let the Zernike basis be defined as,

\begin{equation}
V_n^{m}(r,\theta) = R_n^{|m|}(r)\,e^{\mathrm{i}m\theta},
\end{equation}

where $R_n^{|m|}(r)$ is the radial polynomial and $(r,\theta)$ are polar coordinates with $r \in [0,1]$, $\theta \in [0,2\pi)$. We aim to prove,
\begin{equation}
\int_{0}^{1}\!\!\int_{0}^{2\pi} V_n^{m}(r,\theta)\,\big(V_{n'}^{m'}(r,\theta)\big)^{\!*}\,r\,d\theta\,dr
=\frac{\pi}{n+1}\,\delta_{nn'}\delta_{mm'}
\end{equation}

The integrand in the above equation can be separated as,

\begin{equation}
\int_{0}^{1}\!\!\int_{0}^{2\pi} V_n^{m}(r,\theta)\,\big(V_{n'}^{m'}(r,\theta)\big)^{\!*}\,r\,d\theta\,dr
=\left( \int_{0}^{2\pi} e^{\mathrm{i}(m-m')\theta}\,d\theta \right)
\left( \int_{0}^{1} R_n^{|m|}(r)\,R_{n'}^{|m'|}(r)\,r\,dr \right).
\end{equation}

The angular integral yields,

\begin{equation}
\int_{0}^{2\pi} e^{\mathrm{i}(m-m')\theta}\,d\theta = 2\pi\,\delta_{mm'}
\end{equation}

Hence it remains to show that, for fixed $m \ge 0$,

\begin{equation}
\int_{0}^{1} R_n^{m}(r)\,R_{n'}^{m}(r)\,r\,dr = \frac{\delta_{nn'}}{2(n+1)}
\end{equation}

Let $k = \tfrac{n-m}{2} \in \mathbb{N}_0$. A standard identity for Zernike radial polynomials is

\begin{equation}
R_{m+2k}^{\,m}(r) = r^{m}\,P_{k}^{(0,m)}\!\left(2r^{2}-1\right),
\end{equation}

where $P_k^{(\alpha,\beta)}$ is the Jacobi polynomial. Set $x = 2r^{2} - 1$, so that $r\,dr = \frac{1}{4}\,dx$ and $r^{2} = \frac{1+x}{2}$. The radial integral becomes
\begin{equation}
\begin{aligned}
\int_{0}^{1} R_{m+2k}^{\,m}(r)\,R_{m+2k'}^{\,m}(r)\,r\,dr
&= \frac{1}{4}\int_{-1}^{1} \left(\frac{1+x}{2}\right)^{m} P_{k}^{(0,m)}(x) P_{k'}^{(0,m)}(x)\,dx \\
&= 2^{-(m+2)} \int_{-1}^{1} (1+x)^{m} P_{k}^{(0,m)}(x) P_{k'}^{(0,m)}(x)\,dx.
\end{aligned}
\end{equation}

The Jacobi polynomial orthogonality relation (for $\alpha=0$, $\beta=m$) is
\begin{equation}
\int_{-1}^{1} (1-x)^{0}(1+x)^{m} P_{k}^{(0,m)}(x) P_{k'}^{(0,m)}(x)\,dx
= \frac{2^{m+1}}{2k + m + 1} \,\delta_{kk'}
\end{equation}
Therefore,
\begin{equation}
\int_{0}^{1} R_{m+2k}^{\,m}(r)\,R_{m+2k'}^{\,m}(r)\,r\,dr
= 2^{-(m+2)} \cdot \frac{2^{m+1}}{2k + m + 1} \,\delta_{kk'}
= \frac{1}{2(2k + m + 1)}\,\delta_{kk'}
\end{equation}

Since $n = m + 2k$, we have $2k + m + 1 = n + 1$, giving
\begin{equation}
\int_{0}^{1} R_{n}^{\,m}(r)\,R_{n'}^{\,m}(r)\,r\,dr
= \frac{\delta_{nn'}}{2(n+1)}
\end{equation}

Combining with the angular part,

\begin{equation}
\int_{0}^{1}\!\!\int_{0}^{2\pi} V_n^{m}(r,\theta)\,\big(V_{n'}^{m'}(r,\theta)\big)^{\!*}\,r\,d\theta\,dr
= \left( 2\pi\,\delta_{mm'} \right) \cdot \left( \frac{\delta_{nn'}}{2(n+1)} \right)
= \frac{\pi}{n+1} \,\delta_{nn'}\delta_{mm'}.
\end{equation}

This completes the proof. In particular, if we define the normalized basis
\begin{equation}
\widetilde{V}_n^{m}(r,\theta) = \sqrt{\frac{n+1}{\pi}}\,R_n^{|m|}(r)\,e^{\mathrm{i}m\theta},
\end{equation}

then $\{\widetilde{V}_n^{m}\}$ forms an orthonormal set on the unit disk with respect to the measure $r\,dr\,d\theta$.
\end{proof}

\subsection{Zernike Basis Shape Linearity Proof}
\label{sec:app:zernike_linearity_proof}

Let \( f_1(r, \theta) \) and \( f_2(r, \theta) \) be two functions defined over the unit disk \( r \in [0, 1], \theta \in [0, 2\pi) \), and let \( f(r, \theta) = a f_1(r, \theta) + b f_2(r, \theta) \), where \( a, b \in \mathbb{R} \). 

The Zernike moment of order \( (n, m) \) of a function \( f \) is defined as:
\begin{equation}
Z_n^m(f) = \int_0^1 \int_0^{2\pi} f(r, \theta) \, V_n^m(r, \theta)^* \, r \, d\theta \, dr
\end{equation}
where \( V_n^m(r, \theta) \) is the complex-valued Zernike basis function and \( ^* \) denotes complex conjugation.

Start with:
\begin{align}
Z_n^m(f) &= \int_0^1 \int_0^{2\pi} \left( a f_1(r, \theta) + b f_2(r, \theta) \right) V_n^m(r, \theta)^* \, r \, d\theta \, dr \\
&= a \int_0^1 \int_0^{2\pi} f_1(r, \theta) V_n^m(r, \theta)^* \, r \, d\theta \, dr + b \int_0^1 \int_0^{2\pi} f_2(r, \theta) V_n^m(r, \theta)^* \, r \, d\theta \, dr \\
&= a Z_n^m(f_1) + b Z_n^m(f_2)
\end{align}

Therefore, Zernike moments are linear operators:
\begin{equation}
Z_n^m\left( a f_1 + b f_2 \right) = a Z_n^m(f_1) + b Z_n^m(f_2)
\end{equation}

\subsection{Radial Frequency Propagation Linearity Proof}
\label{app:rfreqprop_linearity_proof}

For any complex number $u \in \mathbb{C}$, the polar-form identity
\begin{equation}
|u| e^{i \arg(u)} = u
\end{equation}

\noindent holds, with the natural convention that the expression evaluates to zero when $u=0$. Therefore, the propagation rule in Eqn.~(\ref{eqn:arfreqprop}) can be equivalently rewritten as

\begin{equation}
z_n^m \;\leftarrow\; z_n^m + \lambda\, z_{n-2}^m .
\label{eq:rfreqprop_linear}
\end{equation}

Equation~\eqref{eq:rfreqprop_linear} defines a linear mapping from the coefficient vector
$\{z_{n-2}^m,\, z_n^m\}$ to the updated coefficient $z_n^m$.
The full radial frequency propagation is executed by composing a finite sequence of such linear updates along the radial chain $\{n, n-2, n-4, \ldots\}$. Since a finite composition of linear operators remains linear, the linear frequency propagation is linear on the Zernike coefficient space.

Let $\mathcal{Z}$ denote the Zernike basis transform that maps a real-valued shape geometry field $f$
to its Zernike coefficients $\{z_n^m\}$, $\mathcal{P}_\lambda$ defines the radial frequency propagation operation that is conditioned on the propagation coefficient $\lambda$. From Eqn.~(\ref{eq:rfreqprop_linear}), we can learn that $\mathcal{P}_\lambda$ is a linear operation. The final shape geometry encoding $E_\lambda(f)$ can be represented as,

\begin{equation}
E_\lambda(f) := \mathcal{P}_\lambda\big(\mathcal{Z}(f)\big)
\end{equation}

We can then derive that $E_\lambda$ is a linear functional of the input field $f$.
That is, for any scalars $a,b \in \mathbb{R}$ and any geometry fields $f_1,f_2$,

\begin{equation}
E_\lambda(a f_1 + b f_2)
= a\,E_\lambda(f_1) + b\,E_\lambda(f_2).
\end{equation}

Therefore, the composition $E_\lambda = \mathcal{P}_\lambda \circ \mathcal{Z}$ is linear, which completes the proof.

\subsection{Harmonic Pose Field Derivation}
\label{sec:app:harmonic_posefield_derive}

In this section, we derive the Zernike projection coefficient of a harmonic pose field $f_{\text{pose}}(r, \theta; m_p)$ onto the Zernike basis $V_n^m(r, \theta)$. The complex Zernike basis is defined as,

\begin{equation}
V_n^m(r, \theta) = R_n^{|m|}(r) \cdot e^{i m \theta}
\end{equation}

The harmonic pose field is defined as,

\begin{equation}
f_{\text{pose}}(r, \theta; m_p, \mathbf{p}) = \left( \sum_{k=1}^K p_k w_k(r) \right) \cos(m_p \theta)
\label{eqn:app:pose_field}
\end{equation}

Adopting Euler’s formula: $\cos(m_p \theta) = \frac{1}{2} \left( e^{i m_p \theta} + e^{-i m_p \theta} \right)$, we can rewrite Eqn.~(\ref{eqn:app:pose_field}) as:

\begin{equation}
f_{\text{pose}}(r, \theta; m_p, \mathbf{p}) = \frac{1}{2} \sum_{k=1}^K p_k w_k(r) \left( e^{i m_p \theta} + e^{-i m_p \theta} \right)
\end{equation}

Project $f_{\text{pose}}$ onto $V_n^m$,

\begin{equation}
a_n^m = \int_0^{2\pi} \int_0^1 f_{\text{pose}}(r, \theta; m_p) \cdot \overline{V_n^m(r, \theta)} \cdot r\,dr\,d\theta
\end{equation}

Substitute $f_{\text{pose}}$ and $V_n^m$,
\begin{equation}
a_n^m = \frac{1}{2} \sum_{k=1}^K p_k \int_0^{2\pi} \int_0^1 w_k(r) \left( e^{i m_p \theta} + e^{-i m_p \theta} \right) \cdot R_n^{|m|}(r) e^{-i m \theta} \cdot r\,dr\,d\theta
\end{equation}

Separate integrals,

\begin{equation}
a_n^m = \frac{1}{2} \sum_{k=1}^K p_k \left( \int_0^{2\pi} e^{i (m_p - m) \theta} d\theta + \int_0^{2\pi} e^{-i (m_p + m) \theta} d\theta \right) \cdot \int_0^1 w_k(r) R_n^{|m|}(r) r\,dr
\end{equation}

We now evaluate:
\begin{equation}
\int_0^{2\pi} e^{i a \theta} d\theta =
\begin{cases}
2\pi, & a = 0 \\
0, & a \neq 0
\end{cases}
\label{eqn:app:euler_integral}
\end{equation}

We can learn from Eqn.~(\ref{eqn:app:euler_integral}) that we get non-zero values only when $m = \pm m_p$. So,

\begin{equation}
a_n^m = \pi \sum_{k=1}^K p_k \int_0^1 w_k(r) R_n^{|m|}(r) r\,dr, \quad \text{if } m = \pm m_p
\end{equation}

By further adding the Zernike normalization factor $\sqrt{\frac{n+1}{\pi}}$ we can finally get,

\begin{equation}
a_n^m = 
\begin{cases}
\displaystyle
\pi \sqrt{\tfrac{n+1}{\pi}} \sum\limits_{k=1}^K p_k \int_0^1 w_k(r)\, R_n^{|m|}(r)\, r\, dr, & m = \pm m_p \\
0, & \text{otherwise}
\end{cases}
\end{equation}

\subsection{Propagation Ratio $\lambda$ Discussion}
\label{sec:app:prop_ratio_choice}

We enrich higher radial orders within each angular harmonic by cascading from the lower neighbor while preserving rotation behavior. The update is
\begin{equation}
    z_n^m \leftarrow z_n^m + \lambda |{z}_{n-2}^{m}|\cdot e^{i\arg(Z_{n-2}^{m})},
    \label{eqn:highfreq_inject_app}
\end{equation}

applied along fixed $m$ with valid $n$ (\textit{e.g.}, $n-|m|$ even). Writing $n=|m|+2k$ and letting $z_k^m$ denote the coefficient at radial index $k$, the cascaded forward map admits the closed form
\begin{equation}
    z_{k}^{m,\text{mod}}=\sum_{j=0}^{k}\lambda^{j}\,z_{k-j}^m,
    \qquad k=0,1,\dots
\end{equation}
and is exactly invertible via the nilpotent shift operator $S$ along $k$:
\begin{equation}
    \mathbf z^{m}=(I-\lambda S)\,\mathbf z^{m,\text{mod}}.
\end{equation}

We set $\lambda=0.6$. This keeps the per-chain gain bounded by
$\|(I-\lambda S)^{-1}\|\le 1/(1-\lambda)=2.5$ (well-conditioned), while providing a controlled high-frequency tail: the contribution propagated $K$ steps decays as $\lambda^{K}$, e.g., $\lambda^{10}\approx 6\times10^{-3}$. Thus $\lambda=0.6$ yields a meaningful spread across higher $n$ without excessive amplification, and the exact demodulation $(I-\lambda S)$ remains numerically stable.

\subsection{Pose Encoding Linearity and Superposition Proof}
\label{sec:app_pose_linear}

Given the Zernike basis $V_n^{m_p}$ and two pose vectors $\mathbf{p_1}$ and $\mathbf{p_2}$, their linearly combined shape pose encoding by two scalars $\alpha$ and $\beta$ can be represented by 

\begin{equation}
a_n^{m_p}(\alpha \mathbf p_1 + \beta \mathbf p_2)
=\left\langle f_{\mathcal P}(\cdot;\alpha \mathbf p_1 + \beta \mathbf p_2),\, V_n^m \right\rangle
\end{equation}

Using the linear construction of the pose field in Eq.~\eqref{eqn:harmonic_pose_field}, we have

\begin{align}
f_{\mathcal P}(r,\theta;m_p, \alpha \mathbf{p_1} + \beta \mathbf{p_2})
&=\big(\sum_{k=1}^K(\alpha\cdot p_{1,k} + \beta p_{2,k})w_k(r)\big)\cdot\cos(m_p\theta) \\
&=\alpha \sum_{k\in 1}^{K} \big(p_{1,k}w_k(r)\big)\cos(m_p\theta) + \beta \sum_{k\in 1}^{K} \big(p_{2,k}w_k(r)\big)\cos(m_p\theta) \\
&=\alpha\cdot f_{\mathcal P}(r,\theta;\mathbf{p_1}) \;+\; \beta\, f_{\mathcal P}(r,\theta;\mathbf{p_2}).
\label{eq:pose_field_additivity}
\end{align}

Substituting Eqn.~\eqref{eq:pose_field_additivity} into the inner product and using linearity of the integral (hence linearity of $\langle \cdot,\cdot\rangle$ in its first argument), we obtain,

\begin{align}
a_n^{m_p}(\alpha \mathbf{p_1} + \beta \mathbf{p_2})
&=\left\langle \alpha f_{\mathcal P}(\cdot;\mathbf p_1) + \beta f_{\mathcal P}(\cdot;\mathbf p_2),\, V_n^{m_p} \right\rangle \\
&=\alpha \left\langle f_{\mathcal P}(\cdot;\mathbf p_1),\, V_n^{m_p} \right\rangle
\;+\;
\beta \left\langle f_{\mathcal P}(\cdot;\mathbf p_2),\, V_n^{m_p} \right\rangle \\
&=\alpha\,a_n^{m_p}(\mathbf p_1) + \beta\,a_n^{m_p}(\mathbf p_2).
\end{align}

Thus, the linearity and superposition properties of shape pose encoding hold.